\documentclass{article}

\PassOptionsToPackage{numbers, compress}{natbib}
\usepackage[final]{neurips_2022}

\usepackage{amssymb}
\usepackage{abbr}
\usepackage{times}
\usepackage{epsfig}
\usepackage{graphicx}
\usepackage{amsmath}

\usepackage{float}
\usepackage{subcaption}
\usepackage{indentfirst} 
\usepackage{booktabs}
\usepackage[ruled,linesnumbered, boxed]{algorithm2e}
\usepackage{multirow}
\usepackage{diagbox}
\usepackage{makecell}
\usepackage{appendix}
\usepackage{colortbl}
\definecolor{mygray}{gray}{.9}
\usepackage[pagebackref=true,breaklinks=true,letterpaper=true,colorlinks,bookmarks=false]{hyperref}

\title{Learning Adversarial Semantic Embeddings for Zero-Shot Recognition in Open Worlds}

\newcommand*\samethanks[1][\value{footnote}]{\footnotemark[#1]}

\author{
  Tianqi Li$^1$\\
  \texttt{tianqili@buaa.edu.cn}  
  \And
  Guansong Pang$^2$\thanks{Corresponding author: G. Pang (gspang@smu.edu.sg) and X. Bai (baixiao@buaa.edu.cn)}\\
  \texttt{gspang@smu.edu.sg} \\
  \And
  Xiao Bai$^1$\samethanks\\
  \texttt{baixiao@buaa.edu.cn} \\
  \And
  Jin Zheng$^1$\\
  \texttt{JinZheng@buaa.edu.cn} \\
  \And
  Lei Zhou$^3$\\
  \texttt{leizhou@whut.edu.cn} \\
  \And
  Xin Ning$^4$\\
  \texttt{ningxin@semi.ac.cn} \\
  \AND \\
  $^1$ School of Computer Science and Engineering, Beihang University \\
  $^2$ School of Computing and Information Systems, Singapore Management University\\
  $^3$ School of Computer Science and Artificial Intelligence, Wuhan University of Technology \\
  $^4$ Institute of Semiconductors, Chinese Academy of Sciences
}
\begin{document}

\maketitle

\begin{abstract}
  Zero-Shot Learning (ZSL) focuses on classifying samples of \textbf{unseen classes} with only their side semantic information presented during training. It cannot handle real-life, open-world scenarios where there are test samples of \textbf{unknown classes} for which neither samples (\eg, images) nor their side semantic information is known during training. Open-Set Recognition (OSR) is dedicated to addressing the unknown class issue, but existing OSR methods are not designed to model the semantic information of the unseen classes. To tackle this combined ZSL and OSR problem, we consider the case of ``Zero-Shot Open-Set Recognition" (ZS-OSR), where a model is trained under the ZSL setting but it is required to accurately classify samples from the unseen classes while being able to reject samples from the unknown classes during inference. We perform large experiments on combining existing state-of-the-art ZSL and OSR models for the ZS-OSR task on four widely used datasets adapted from the ZSL task, and reveal that ZS-OSR is a non-trivial task as the simply combined solutions perform badly in distinguishing the unseen-class and unknown-class samples.
We further introduce a novel approach specifically designed for ZS-OSR, in which our model learns to generate adversarial semantic embeddings of the unknown classes to train an unknowns-informed ZS-OSR classifier. Extensive empirical results show that our method 1) substantially outperforms the combined solutions in detecting the unknown classes while retaining the classification accuracy on the unseen classes and 2) achieves similar superiority under generalized ZS-OSR settings. Our code is available at \href{https://github.com/lhrst/ASE}{https://github.com/lhrst/ASE}.
\end{abstract}

\section{Introduction}
\textit{'A zebra is a horse with black and white striped coats.'} With this description, a child who has never seen a zebra can recognize it at first sight. Humans can recognize images of such unseen classes using the shared semantic knowledge learned from images of the classes previously seen. Inspired by this phenomenon, Zero-Shot Learning (ZSL) was proposed 
to learn a multi-modal cognition ability in image classification~\cite{ZSL}. Given only some side semantic information like attribute vectors or description text of a set of targeted classes (\textbf{unseen classes}), together with samples (\eg, images) of another set of classes (\textbf{seen classes}) and their semantic information, ZSL aims to learn a model to recognize images of the unseen classes. Many ZSL methods have been introduced over the years, including embedding methods~\cite{Xu_2022_CVPR, Chen_Hong_Liu_Xie_Sun_Li_Peng_Lu_You_2022} that learn mappings between the given semantic knowledge and images in a new feature space, and generative methods~\cite{Verma_2018_CVPR,xian2019f, narayan2020latent} that train a generator to synthesize training samples for the unseen classes, transforming the ZSL to a standard image recognition task.

\begin{figure}
  \centering
  \includegraphics[width=0.75\linewidth]{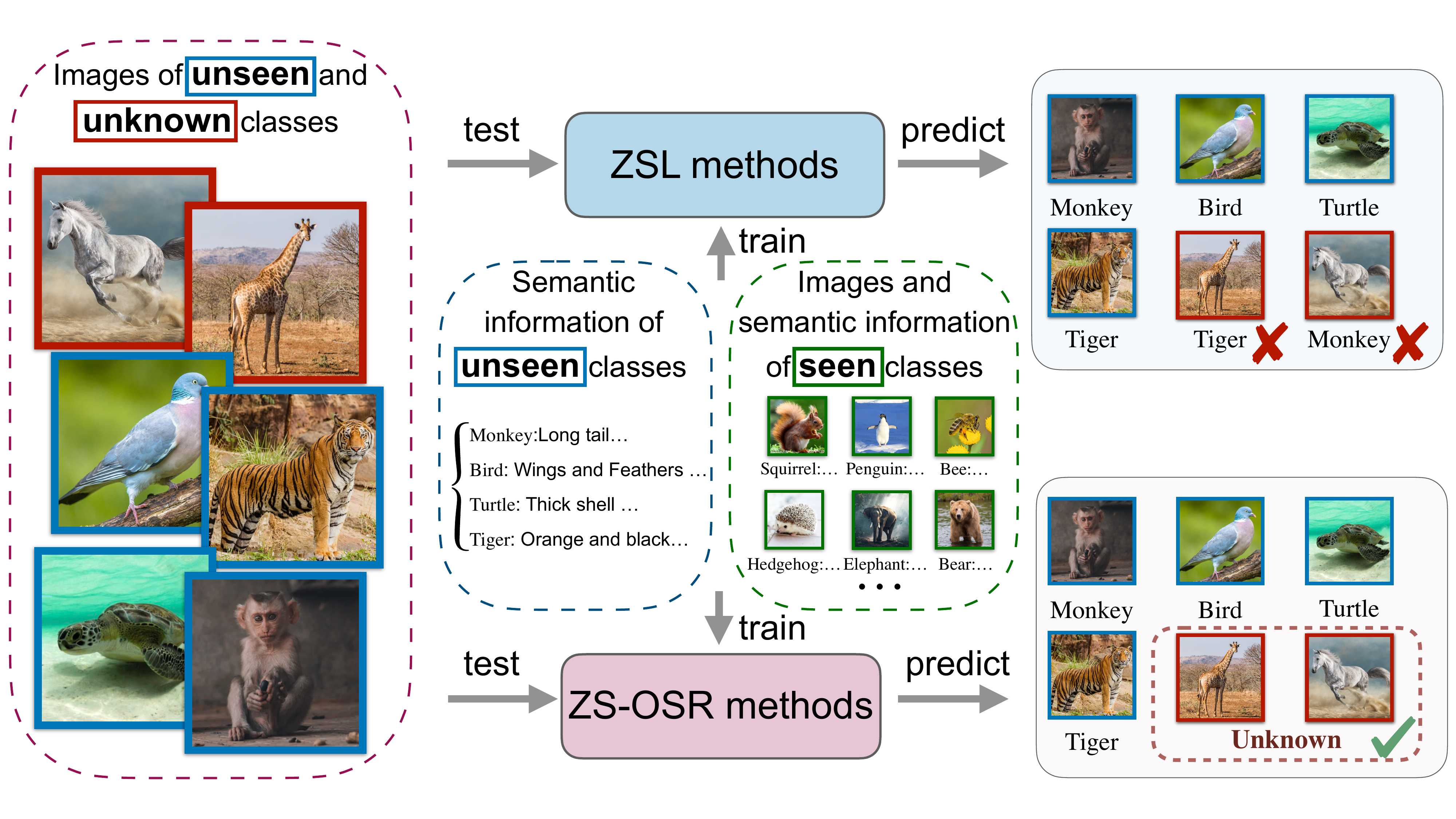}
  \caption{The ZS-OSR task vs. the ZSL task. 
  ZS-OSR considers open-set scenarios where test data can also contain samples of unknown classes, unlike the closed-set setting in ZSL that presumes the presence of unseen-class samples only. As a result, ZSL methods would misclassify the unknown-class samples into one unseen class, whereas ZS-OSR methods can reject these unknown samples while correctly classifying the unseen-class samples.}
  \label{fig:motivation_fig}
\end{figure}

However, current ZSL approaches cannot handle real-life, open-world scenarios, where there are test samples of \textbf{unknown classes} for which neither samples (\eg, images) nor their side semantic information is known during training, such as unknown objects encountered in self-driving contexts and chest X-ray images of unknown viral pneumonia. This is because they assume a closed-set learning setting where all classes in the test data are presented in the training data. As a result, ZSL methods would misclassify the unknown-class samples into one of the unseen classes, as demonstrated in Figure \ref{fig:motivation_fig} (top). Open-Set Recognition (OSR)~\cite{6365193,geng2020recent} is dedicated to addressing the unknown class issue, but existing OSR methods are not designed to model the semantic information of the unseen classes. 
To tackle this problem, we consider the case of ``\textit{Zero-Shot Open-Set Recognition}" (ZS-OSR), where a model is trained under the ZSL setting but it is required to accurately classify samples from the unseen classes while being able to reject samples from the unknown classes during inference, as illustrated in Figure \ref{fig:motivation_fig} (bottom). As shown in Table \ref{tab:setting}, ZS-OSR is a joint ZSL and OSR problem. A straightforward solution is thus to simply combine existing state-of-the-art (SOTA) ZSL and OSR models to build ZS-OSR models. As a contribution to Pattern Recognition research, we establish a set of such baselines and construct four ZS-OSR benchmark datasets adapted from widely-used ZSL datasets, \ie, CUB~\cite{WahCUB_200_2011}, AWA2~\cite{8413121}, FLO~\cite{4756141} and SUN~\cite{6247998}, to evaluate their performance. Our empirical results reveal that such combined solutions perform badly in differentiating the unseen-class and the unknown-class samples. These findings highlight the need for more effective solutions for ZS-OSR, which can contribute to the development of more robust and effective pattern recognition systems.

\begin{table}[t]
	\begin{footnotesize}
		\centering
		\begin{tabular}{c|c|c}
            \hline
            \diagbox{Task}{Set} & Training & Testing \\
            \hline
            \textit{ZSL} & \makecell[c]{Seen classes \\ \& $\text{semantic information}^1$} &  \makecell[c]{ Unseen classes}\\
            \hline
			\textit{OSR} &  \makecell[c]{Seen classes
   } & \makecell[c]{Seen classes \\ \& unknown classes}\\
            \hline
			\textit{ZS-OSR} & \makecell[c]{Seen classes \\ \& $\text{semantic information}^1$} & \makecell[c]{Unseen classes \\ \& unknown classes}\\
			\hline
		\end{tabular}
            \vspace{0.3cm}
		\caption{ZS-OSR vs. ZSL and OSR. Here  $\textit{semantic information}^1$ is the semantic information of both seen and unseen classes.}
		\label{tab:setting}
	\end{footnotesize}
\end{table}

We further introduce a novel approach specifically designed for ZS-OSR, namely ASE, which learns to generate \underline{A}dversarial \underline{S}emantic \underline{E}mbeddings of the unknown classes to train an unknowns-informed open-set classifier. The key insight is to generate meaningful samples of both unseen and unknown classes to train the unknowns-informed open-set classifier. Existing generative ZSL models have demonstrated superior performance in generating the samples/features of the unseen classes based on their learned relations between the samples and the semantic embeddings of the seen classes. The key challenge lies in the generation of unknown-class samples. To address this challenge, we introduce the adversarial semantic embedding learning module that learns a set of adversarial semantic embeddings for the unknown classes so that they are tightly distributed around but separable from the unseen-class embeddings (see Figure \ref{fig:rebuttal_visualization} for a visualization of these generated samples). This is achieved by jointly minimizing a distance loss in the \textit{semantic embedding space} that brings the unknown-class embeddings closer to the given unseen-class embeddings, and an adversarial loss in the \textit{feature space} that pulls the corresponding unknown-class feature prototypes away from the unseen-class features generated by an off-the-shelf trained generative ZSL model. Given the learned semantic embeddings, samples of the unknown classes are generated using the trained generative ZSL model.

There have been OSR methods that generate adversarial samples to represent the unknown-class samples, \eg,~\cite{ijcai2017p469,Neal_2018_ECCV}. However, unlike ASE that can work on both the semantic and feature spaces, they cannot model the rich semantics in the semantic embedding space as they were primarily designed to work in the single feature space,
largely limiting their performance in the ZS-OSR setting. 

In summary, this work makes the following contributions: 1) we explore 
the ZS-OSR problem
and establish extensive performance benchmarks by building a set of baselines based on the combination of existing SOTA ZSL and OSR models on four adapted ZS-OSR widely-used datasets,
2) we propose a novel generative approach for ZS-OSR, which learns a set of adversarial semantic embeddings to represent the unknown classes to train an unknowns-informed open-set classifier, and 3) large-scale empirical results show that our method ASE substantially outperforms the baselines in detecting the unknown classes without degrading the classification accuracy on the unseen classes, and it also achieves similar superiority on generalized ZS-OSR and ZS-OOD (out-of-distribution) settings.

\section{Related Work}
\noindent \textbf{Zero-shot learning} ~\cite{ZSL} utilizes an additional class semantic embedding set to connect the seen and unseen classes. Current ZSL approaches are either to align the images and semantic embeddings ~\cite{Xu_2022_CVPR, Chen_Hong_Liu_Xie_Sun_Li_Peng_Lu_You_2022, NEURIPS2020_fa2431bf, 9577628, pmlr-v97-zhang19l, liu2022zero, chen2023integrating, kim2022discriminative, zhang2020deep, zhang2022zero, zhang2021plug}, or to generate image features of unseen classes and train a closed-set classifier ~\cite{li2019leveraging, bendre2021generalized,Verma_2018_CVPR,xian2019f, narayan2020latent, shermin2022integrated}. Recently, big vision-language models like CLIP~\cite{pmlr-v139-radford21a} have demonstrated significant potential in the realm of ZSL. Nevertheless, they are still focused on the standard ZSL setting. New designs are required to utilize such big models to handle unknown samples as neither samples nor side semantic information of the unknown classes is available during ZS-OSR training.

\noindent\textbf{Open-set recognition}~\cite{6365193,geng2020recent} targets the problem of learning a classifier to reject samples of classes that are unseen during training. A large number of OSR methods have been introduced. Some early studies focus on designing new network layers, such as the OpenMax layer~\cite{bendale2016towards}, while most studies are dedicated to generating pseudo unknown-class samples to train open-set classifiers~\cite{Neal_2018_ECCV, DBLP:journals/corr/abs-2103-15086, Kong_2021_ICCV, cevikalp2023anomaly}. The other studies explore new ways of representing the unknown classes, \eg, through prototype mining ~\cite{lu2022pmal}.

\noindent \textbf{Out-of-distribution (OOD) detection} addresses a similar problem as OSR, but focuses on detecting data from a different distribution. For example, ~\cite{hendrycks2017a, liang2018enhancing, 10.5555/3495724.3497526} tackle the problem by exploiting the prediction logits to define OOD scores and reject samples from different datasets, while ~\cite{NEURIPS2021_01894d6f, 9879414} focused on the class-agnostic information in feature space that is not recoverable from logits.

\noindent \textbf{Zero-shot open-set recognition (ZS-OSR)} has not been explored in previous studies, as far as we know. Some related but different explorations are done in ~\cite{yue2021counterfactual, fu2019vocabulary, geng2020guided,gune2019generalized, chen2020boundary,mancini2021open,10.1145/3503161.3548021,Esmaeilpour2022ZeroShotOD}. Particularly, ~\cite{yue2021counterfactual, fu2019vocabulary, geng2020guided} treat ZSL and OSR as two independent problems, meaning that ZSL-oriented methods handle seen and unseen classes only, while OSR-oriented methods handle seen and unknown classes only. Typically, these methods handle both unknown and unseen classes with the same branch in the network, so they cannot be used to solve ZS-OSR tasks. Consequently they do not have the ability to distinguish between unseen and unknown classes. ~\cite{gune2019generalized, chen2020boundary} attempt to utilize OOD detection models to address the problem of generalized ZSL, solving the GZSL problem while neglecting the potential presence of OOD scenarios. ~\cite{mancini2021open} explores a new task, compositional ZSL, different from the conventional ZSL. While conventional ZSL aims to recognize unseen classes, compositional ZSL is designed to explore and recognize unknown combinations of known patterns. Due to the absence of unknown classes in its assumption, it cannot handle ZS-OSR tasks. ~\cite{10.1145/3503161.3548021, Esmaeilpour2022ZeroShotOD} explore the use of a large CLIP model~\cite{pmlr-v139-radford21a} pretrained with extensive auxiliary data for OOD detection without using any training samples, which differs from zero-shot learning as there are no seen classes involved. Additionally, the task of open-set recognition under a few-shot setting is explored in some recent studies \cite{liu2020few,wang2023glocal}, which addresses a different task from ours as we focus on zero-shot settings.

\section{Zero-Shot OSR and Its Challenges}
\subsection{Problem Statement}
In ZS-OSR, there are three different types of classes, including seen classes $\mathcal{Y}^{seen}$, unseen classes $\mathcal{Y}^{unseen}$, and unknown classes $\mathcal{Y}^{unknown}$, where $\mathcal{Y}^{seen} \cap \mathcal{Y}^{unseen} \cap \mathcal{Y}^{unknown} = \emptyset$. Compared to the $\mathcal{Y}^{unknown}$ classes that the training data does not provide any prior information, the classes in $\mathcal{Y}^{seen}$ and $\mathcal{Y}^{unseen}$ are considered as known classes since the training data contains the image samples of the $\mathcal{Y}^{seen}$ classes and some side semantic information of both $\mathcal{Y}^{seen}$ and $\mathcal{Y}^{unseen}$ classes. ZS-OSR aims to accurately recognize the images of unseen classes while rejecting the images of unknown classes based on these training image samples and semantic information.
\begin{figure*}
  \centering
  \begin{subfigure}{0.95\linewidth}
  \centering
    \includegraphics[width=0.24\linewidth]{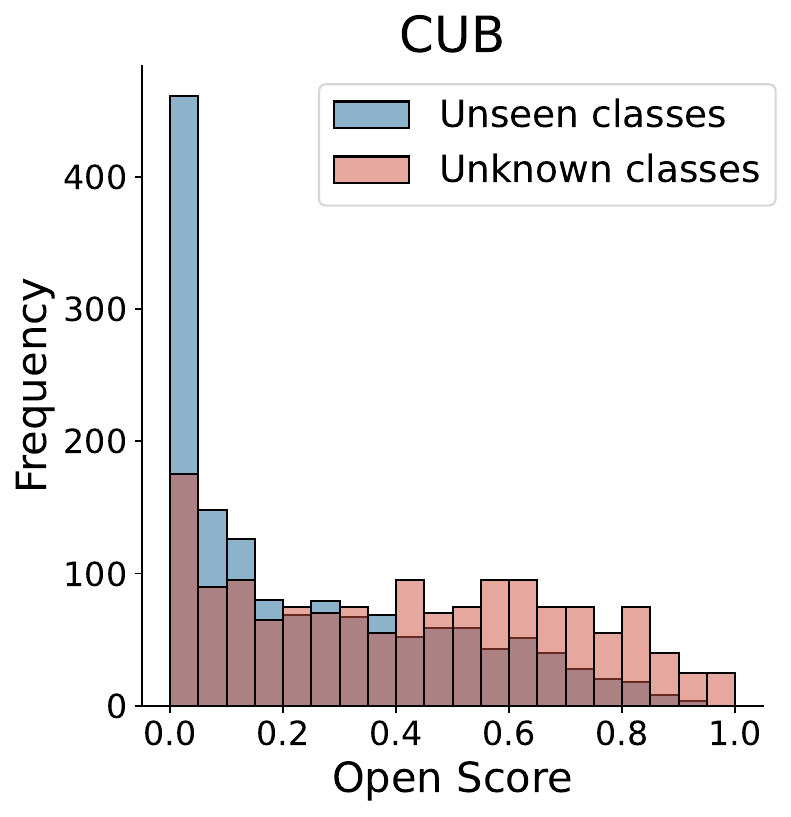}
    \hfill
    \includegraphics[width=0.24\linewidth]{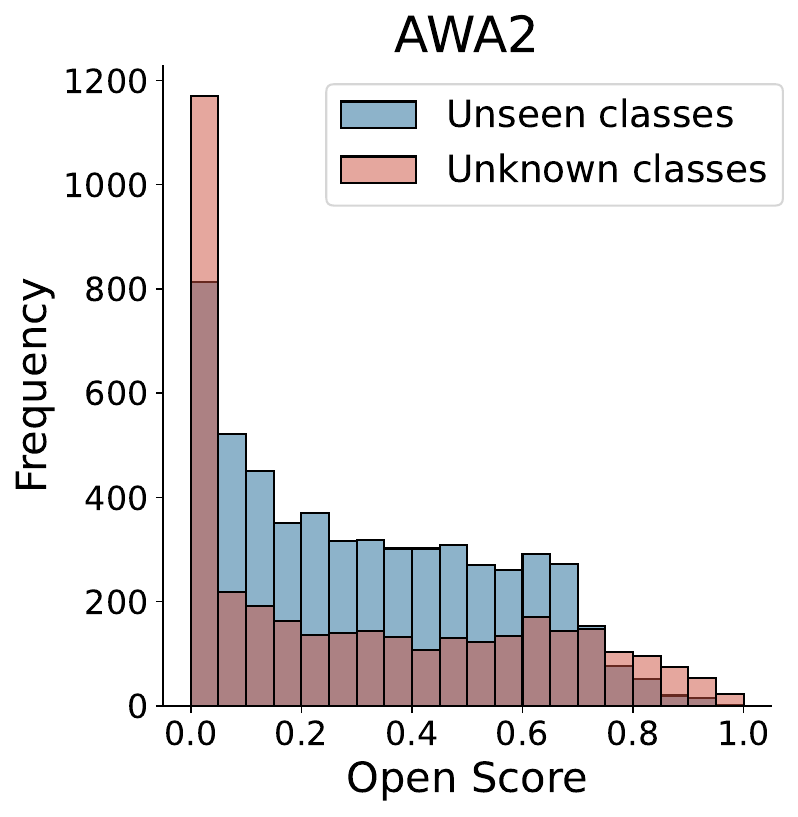}
    \hfill
    \includegraphics[width=0.24\linewidth]{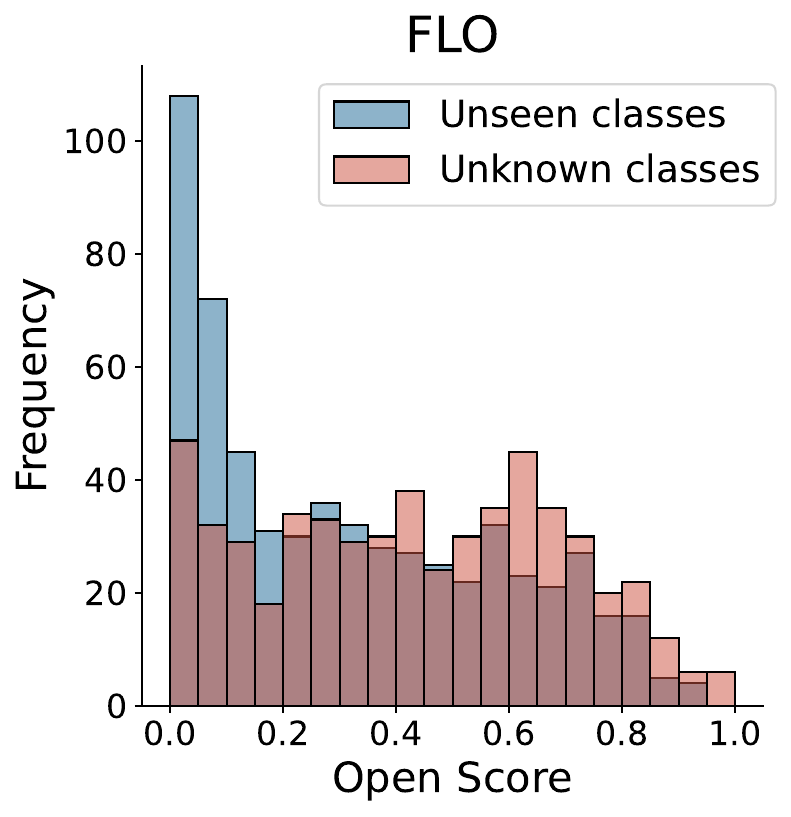}
    \hfill
    \includegraphics[width=0.24\linewidth]{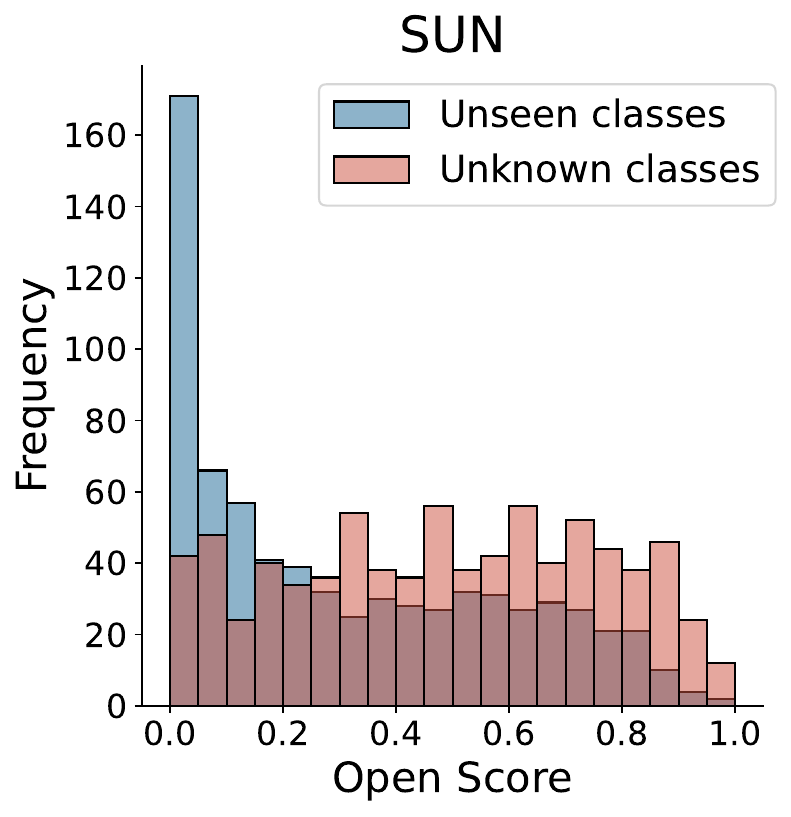}
    \caption{MSP}
  \end{subfigure}
  \begin{subfigure}{0.95\linewidth}
  \centering
    \includegraphics[width=0.24\linewidth]{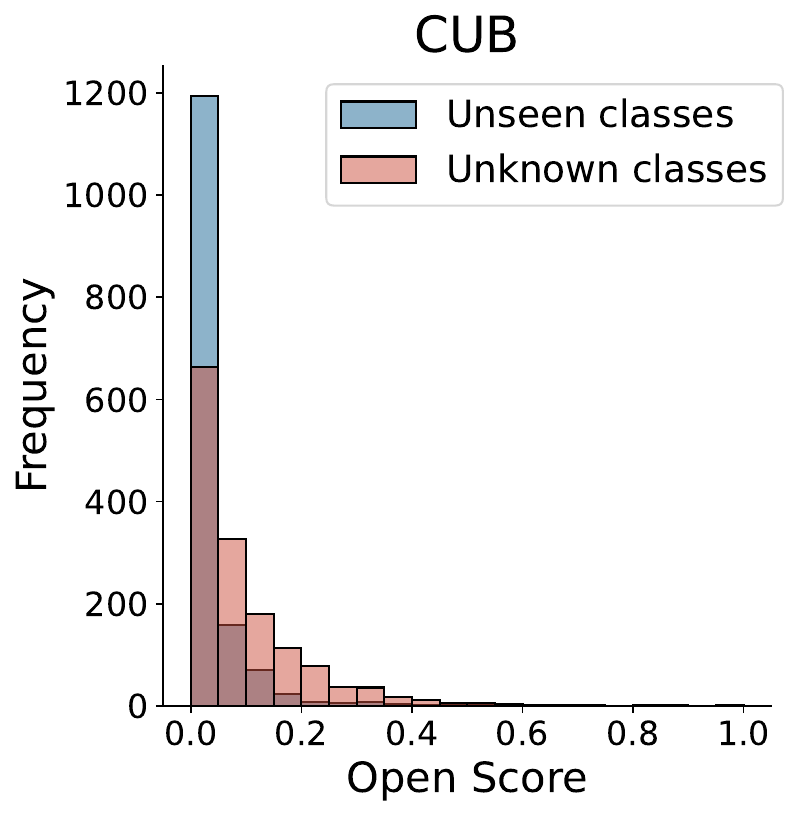}
    \hfill
    \includegraphics[width=0.24\linewidth]{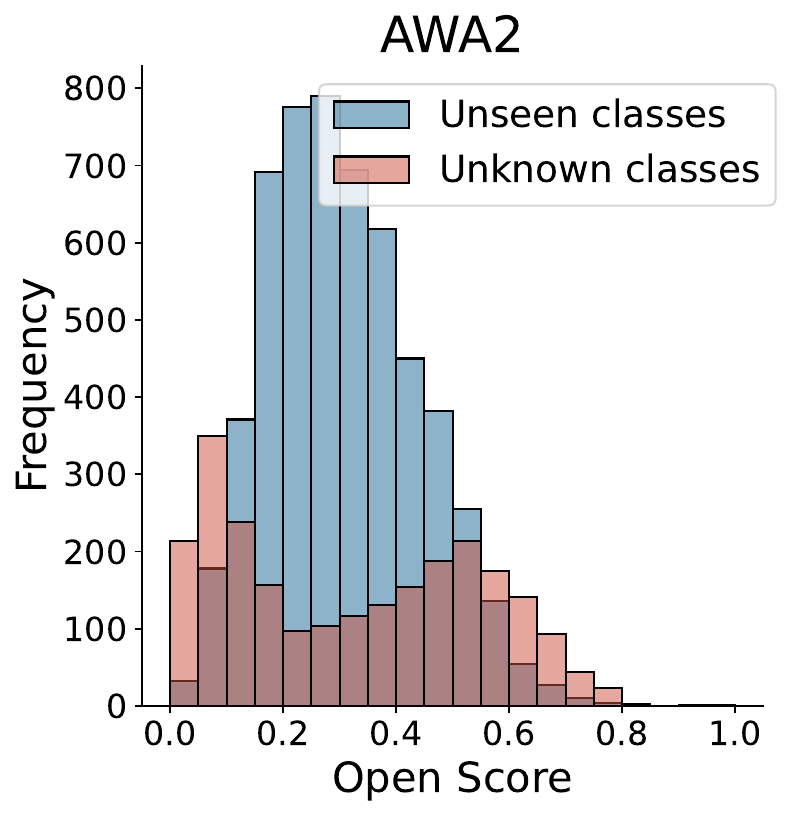}
    \hfill
    \includegraphics[width=0.24\linewidth]{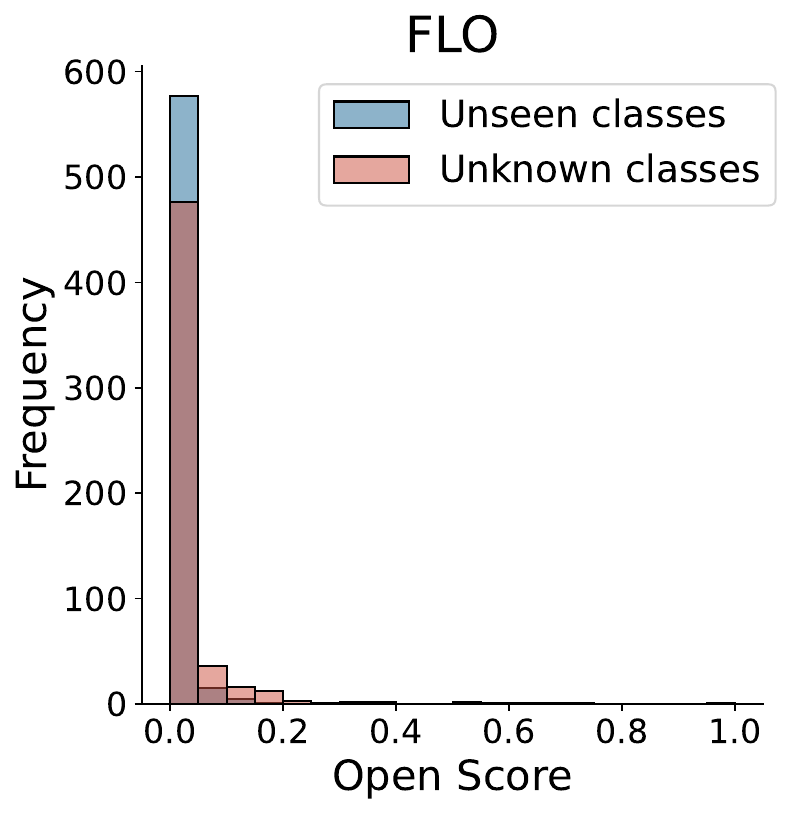}
    \hfill
    \includegraphics[width=0.24\linewidth]{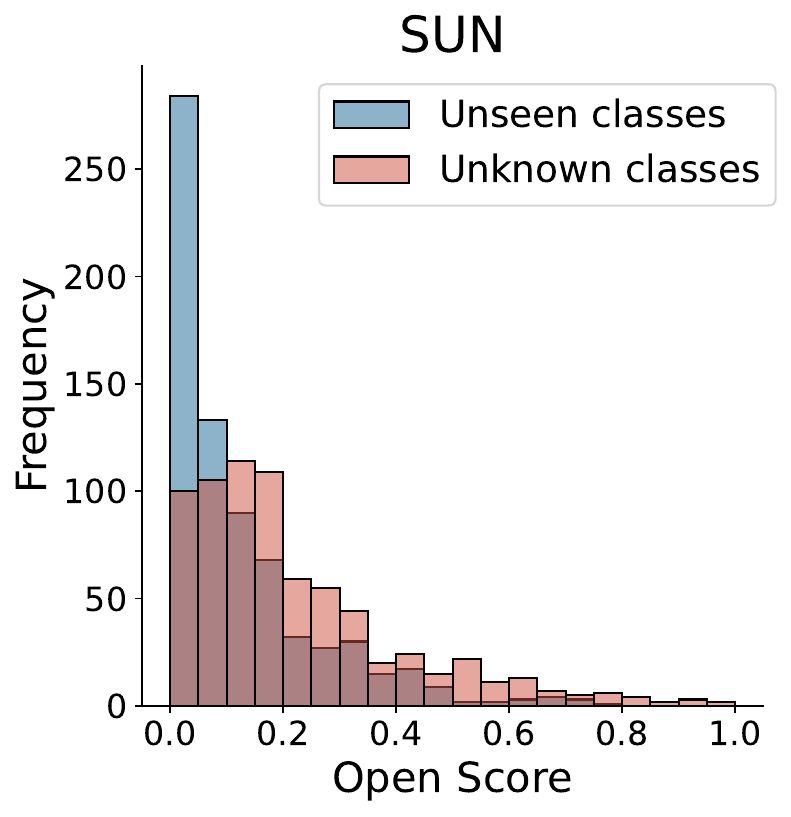}
    \caption{OpenMax}
  \end{subfigure}
  \begin{subfigure}{0.95\linewidth}
  \centering
    \includegraphics[width=0.24\linewidth]{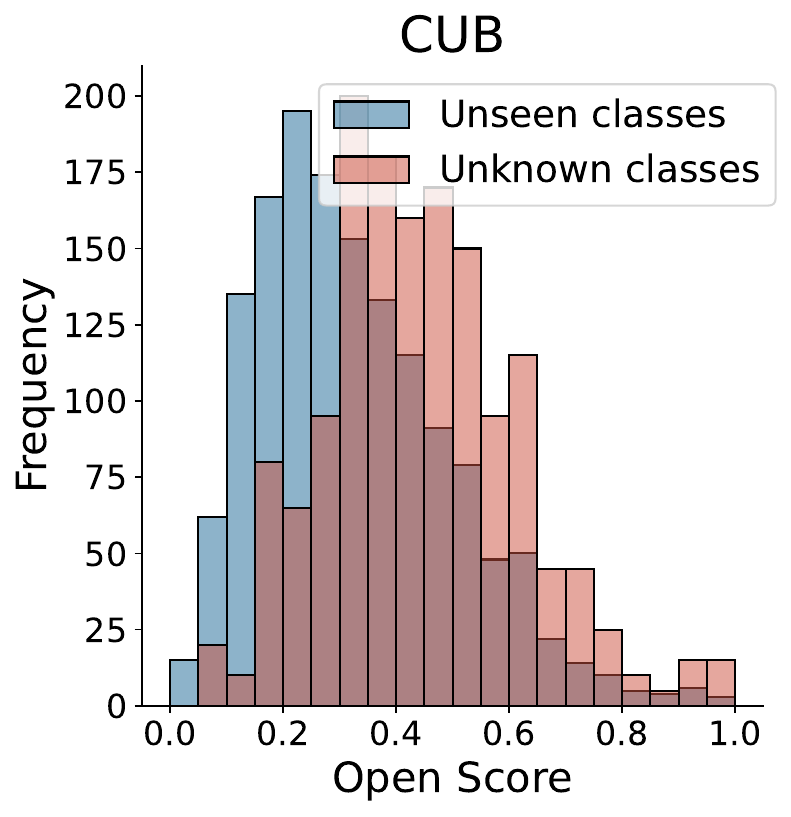}
    \hfill
    \includegraphics[width=0.24\linewidth]{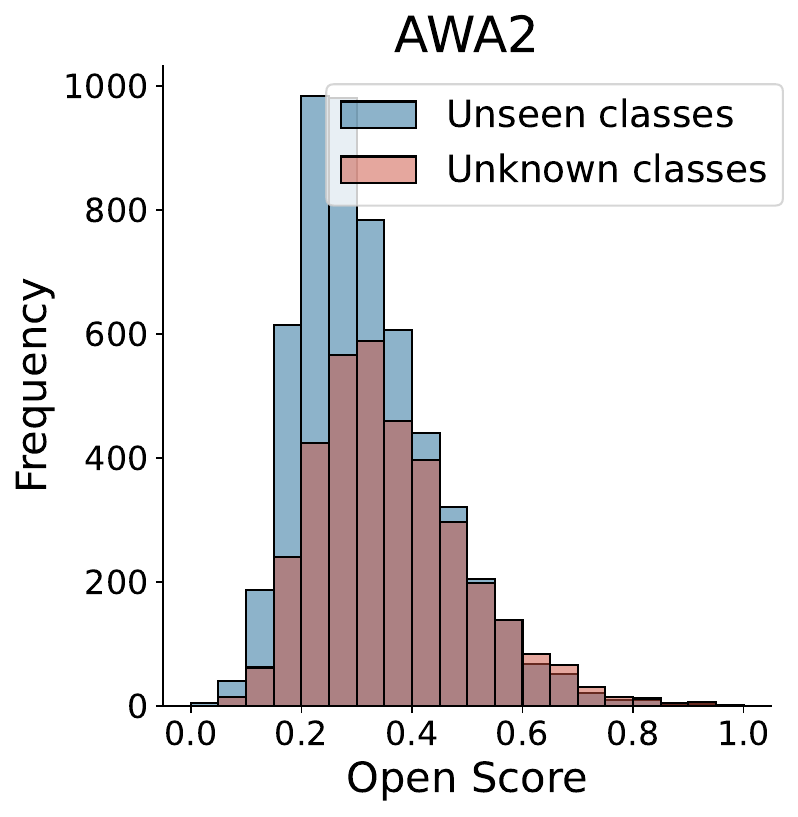}
    \hfill
    \includegraphics[width=0.24\linewidth]{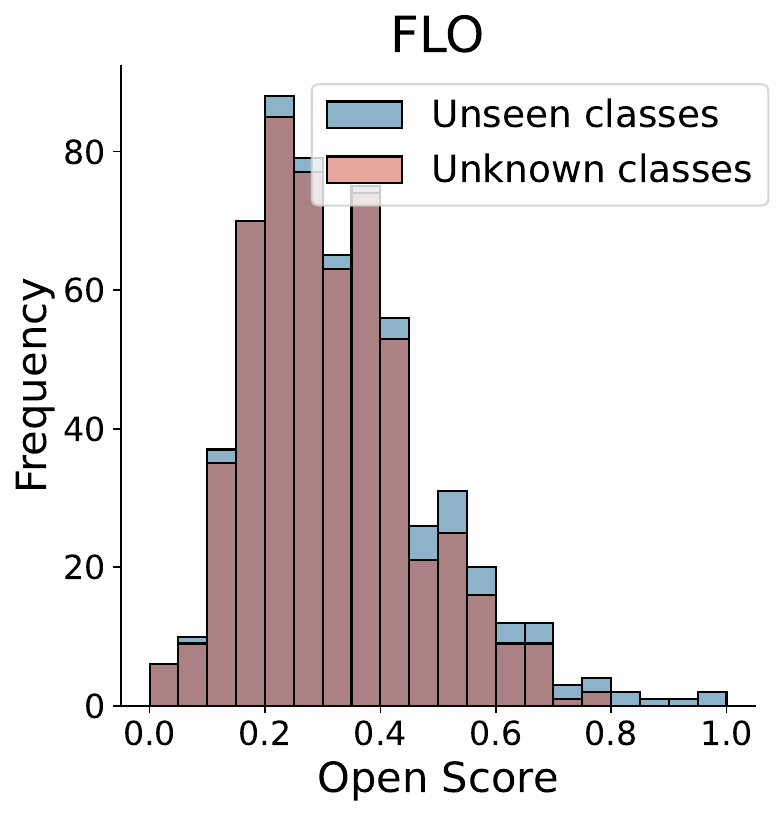}
    \hfill
    \includegraphics[width=0.24\linewidth]{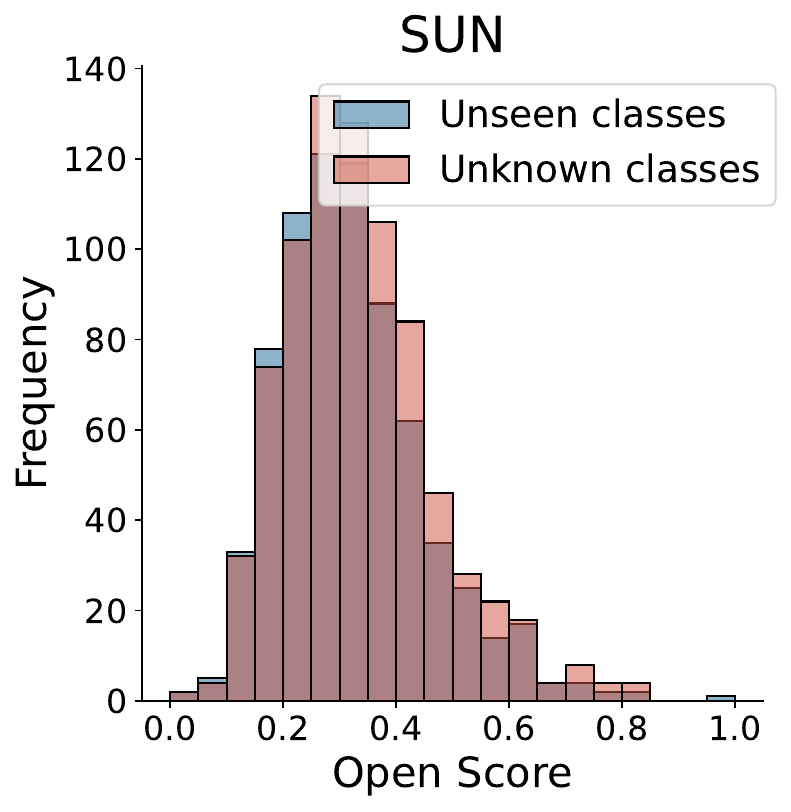}
    \caption{Placeholder}
  \end{subfigure}
  \begin{subfigure}{0.95\linewidth}
  \centering
    \includegraphics[width=0.24\linewidth]{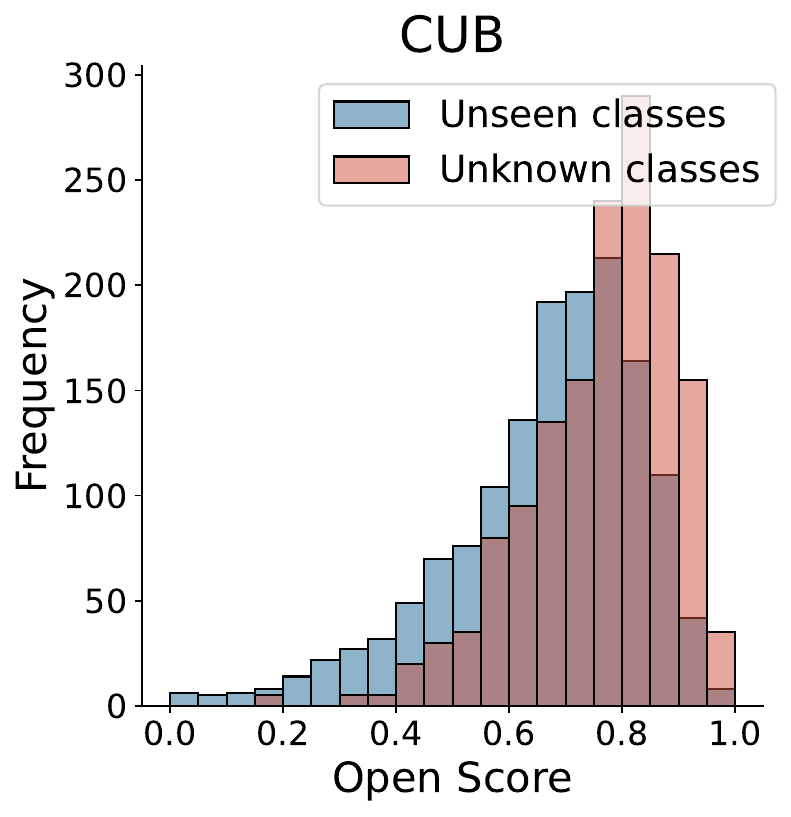}
    \hfill
    \includegraphics[width=0.24\linewidth]{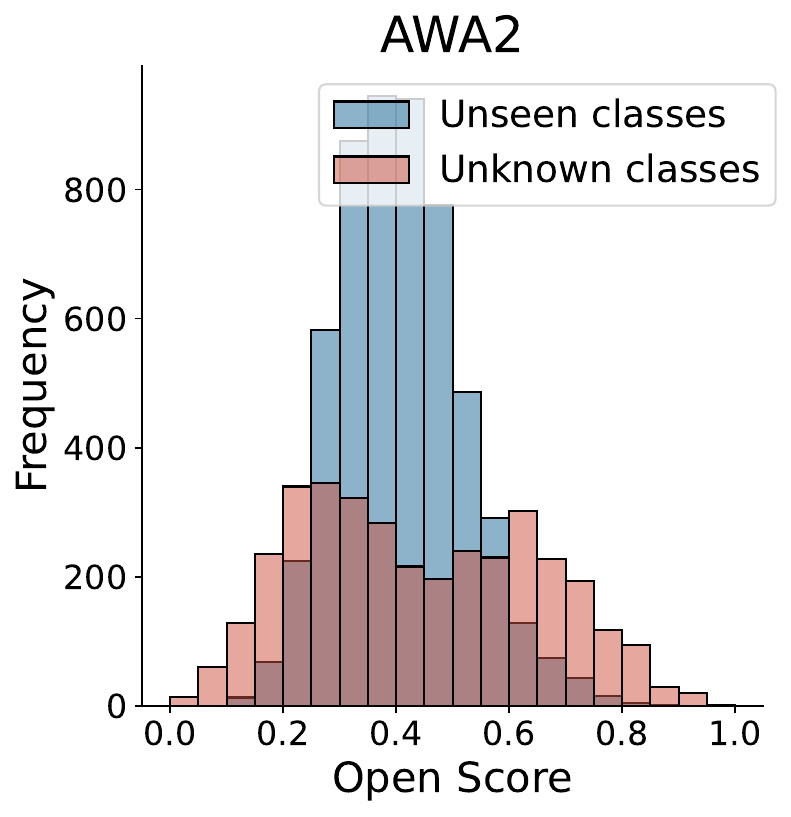}
    \hfill
    \includegraphics[width=0.24\linewidth]{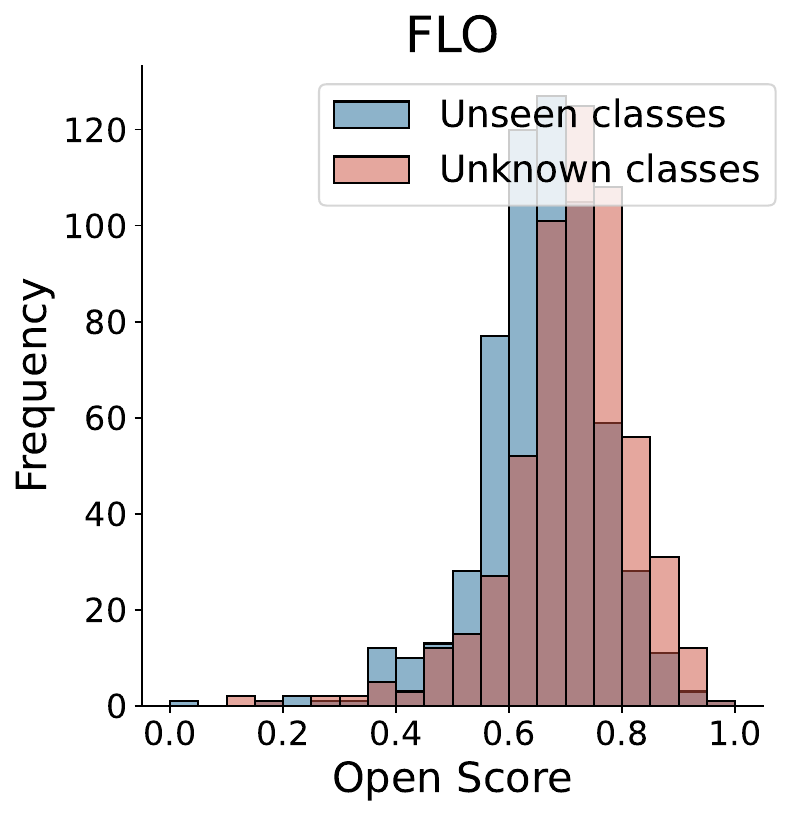}
    \hfill
    \includegraphics[width=0.24\linewidth]{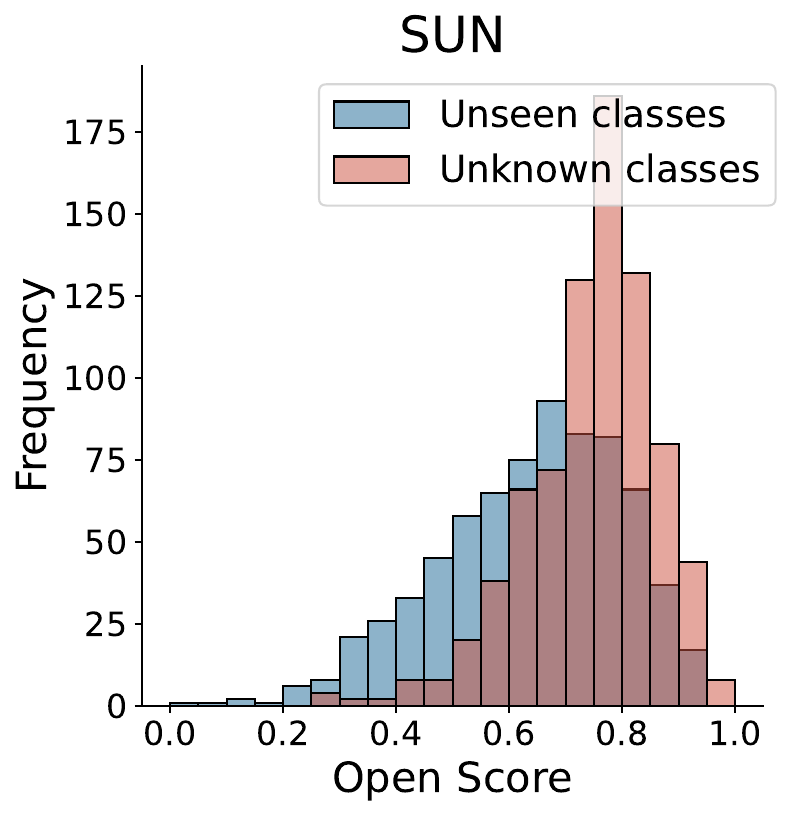}
    \caption{Energy}
  \end{subfigure}
  \begin{subfigure}{0.95\linewidth}
  \centering
    \includegraphics[width=0.24\linewidth]{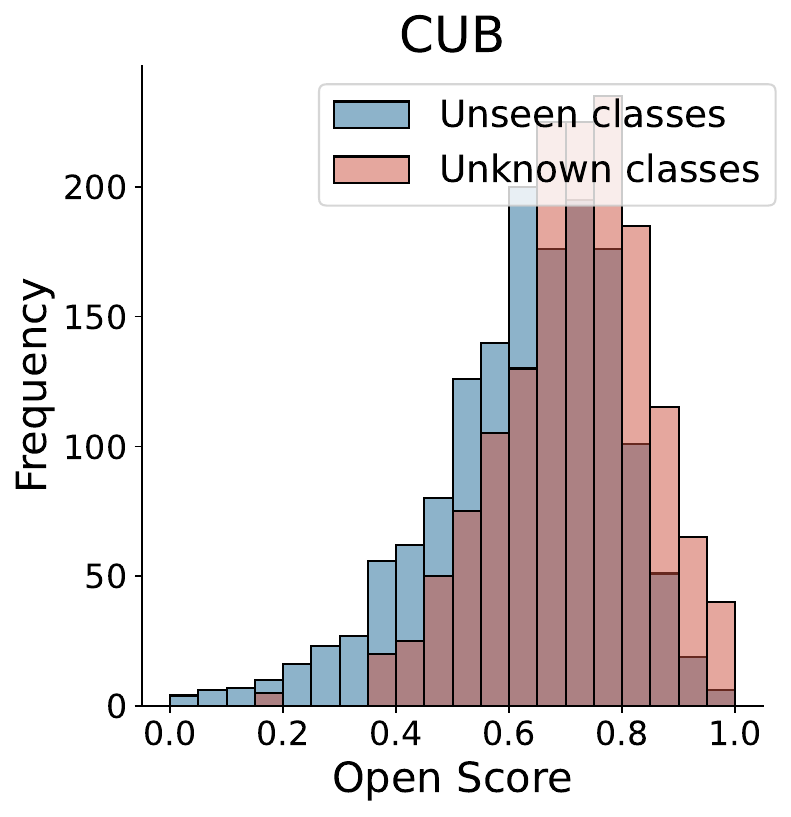}
    \hfill
    \includegraphics[width=0.24\linewidth]{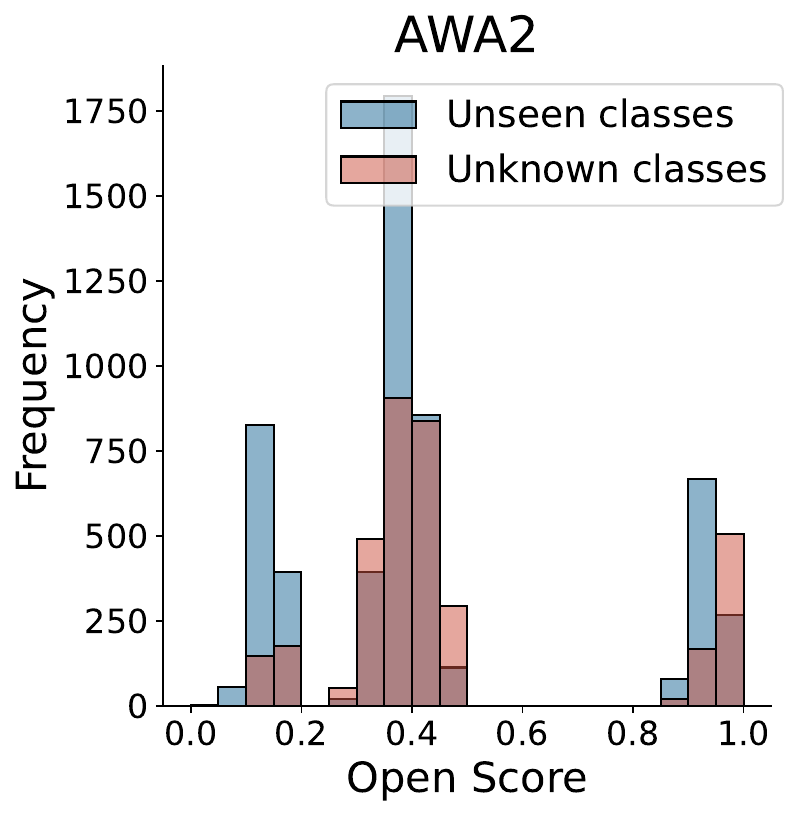}
    \hfill
    \includegraphics[width=0.24\linewidth]{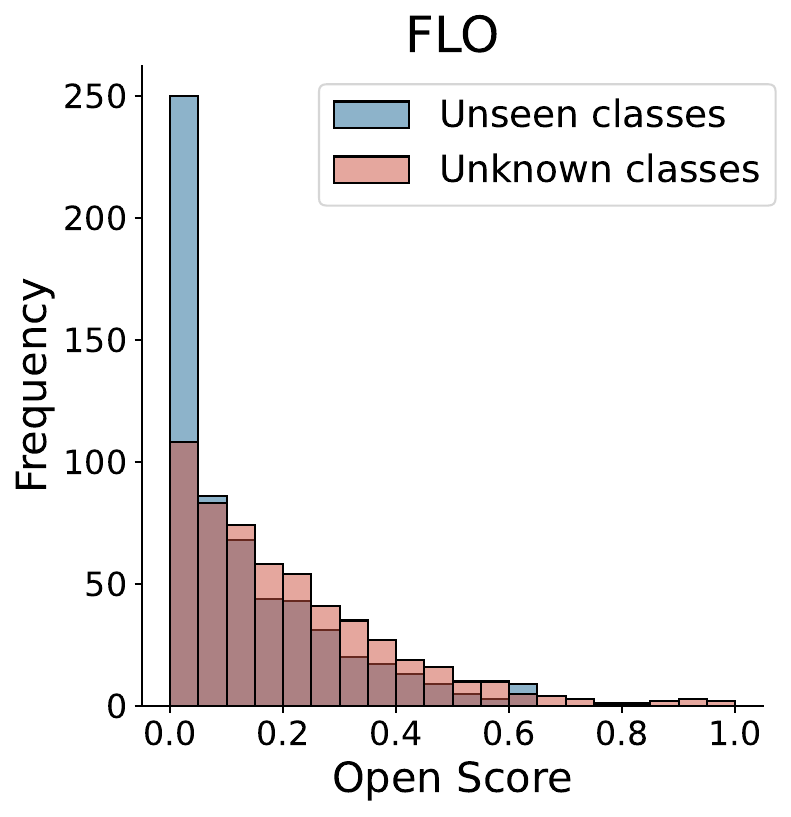}
    \hfill
    \includegraphics[width=0.24\linewidth]{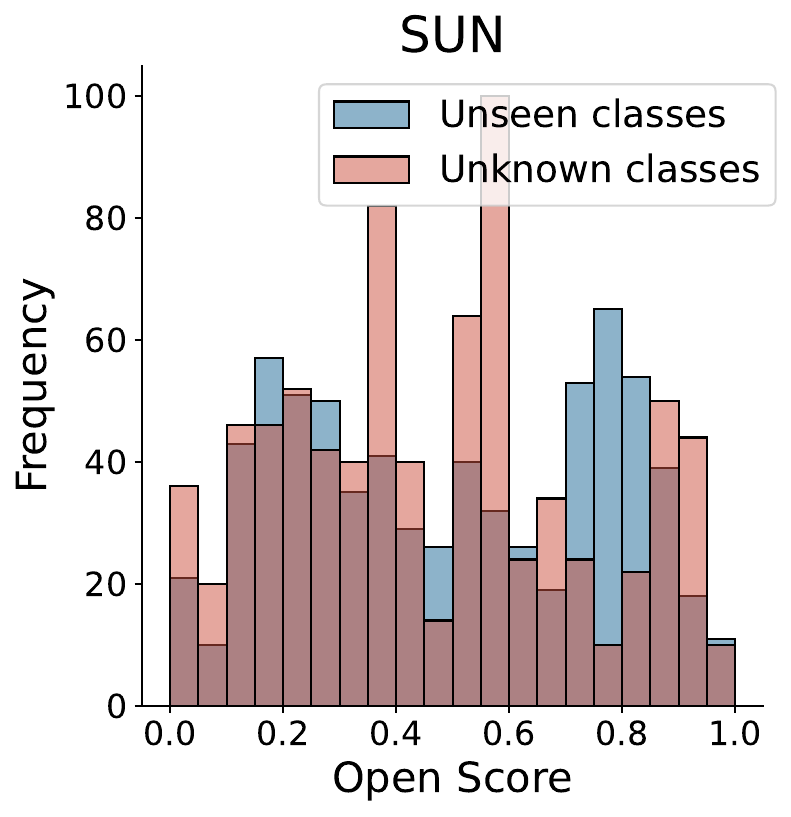}
    \caption{ODIN}
  \end{subfigure}
\end{figure*}
\begin{figure*}
\ContinuedFloat
  \begin{subfigure}{1\linewidth}
  \centering
    \includegraphics[width=0.24\linewidth]{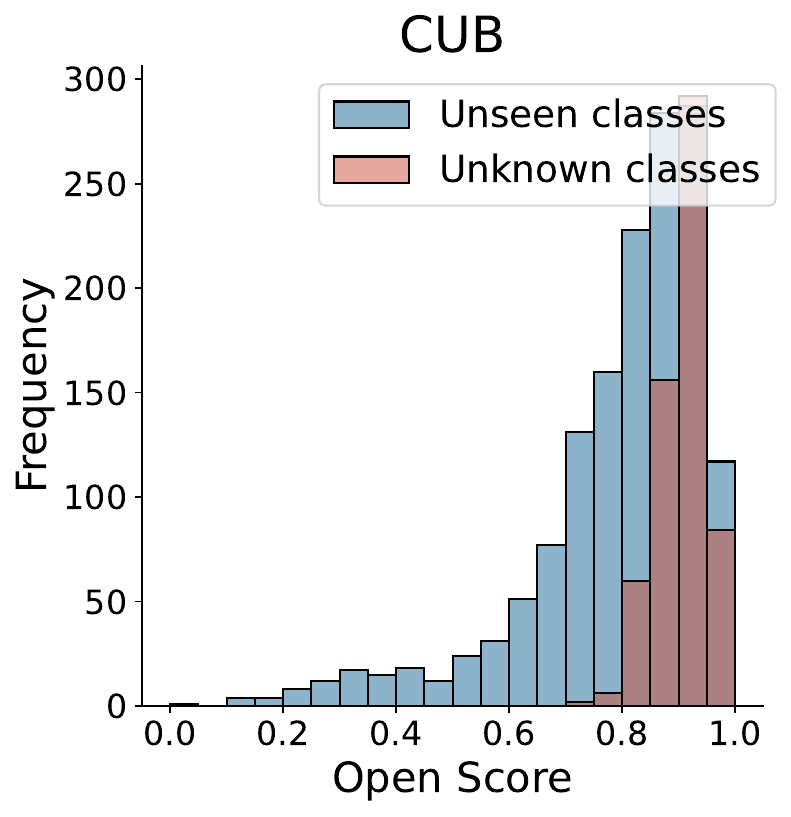}
    \hfill
    \includegraphics[width=0.24\linewidth]{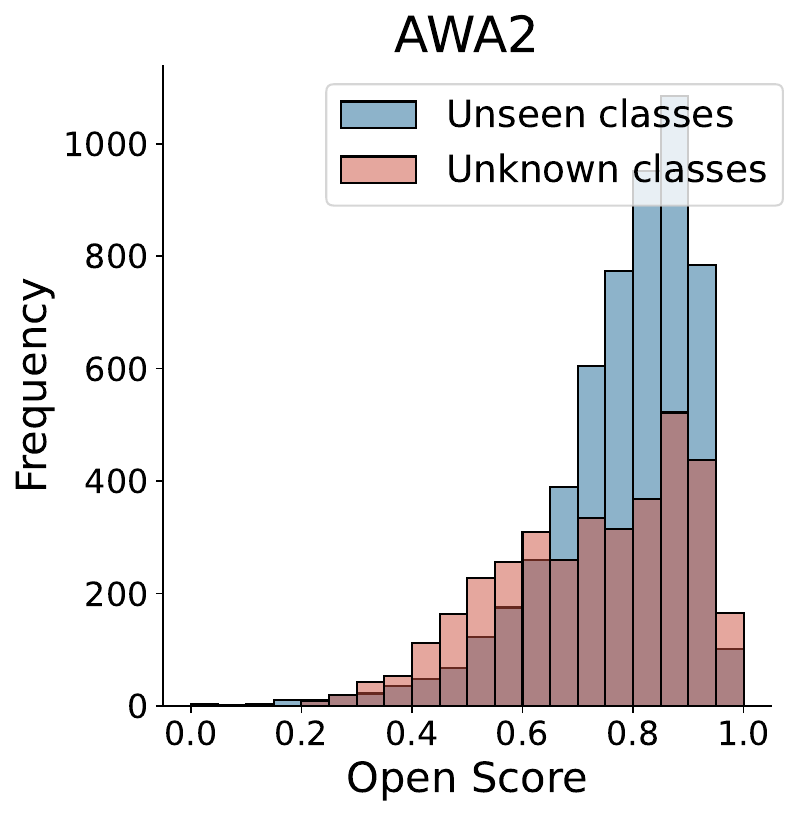}
    \hfill
    \includegraphics[width=0.24\linewidth]{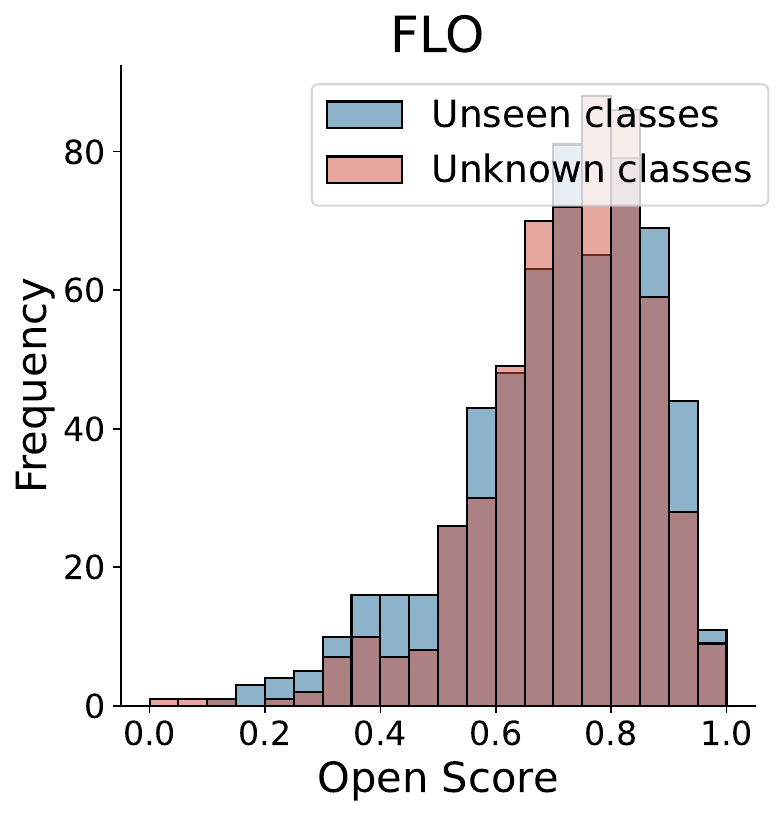}
    \hfill
    \includegraphics[width=0.24\linewidth]{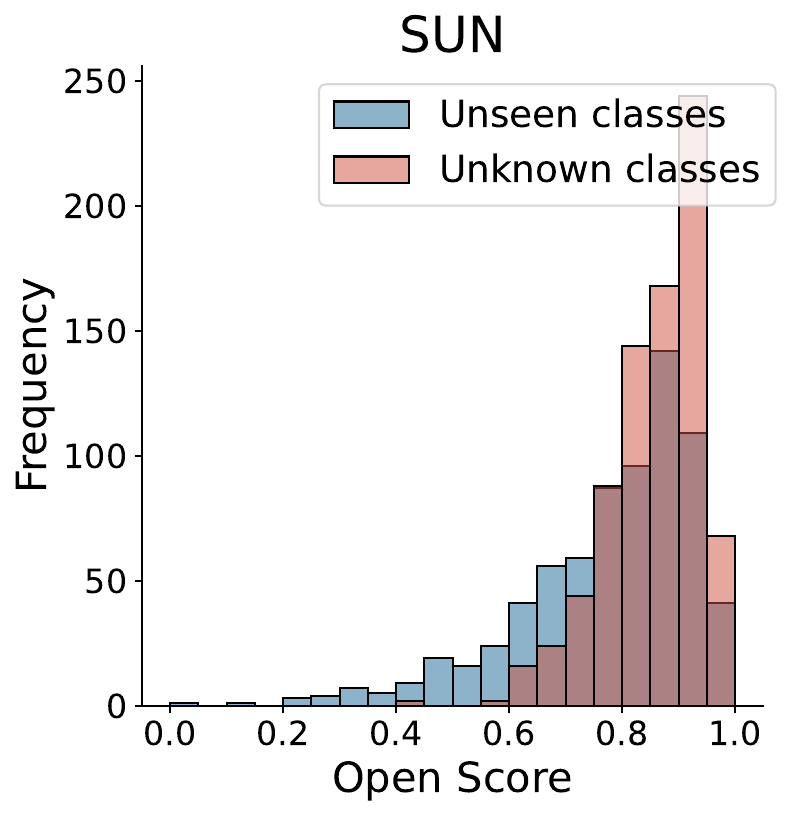}
    \caption{LogitNorm}
  \end{subfigure}
  \begin{subfigure}{1\linewidth}
  \centering
    \includegraphics[width=0.24\linewidth]{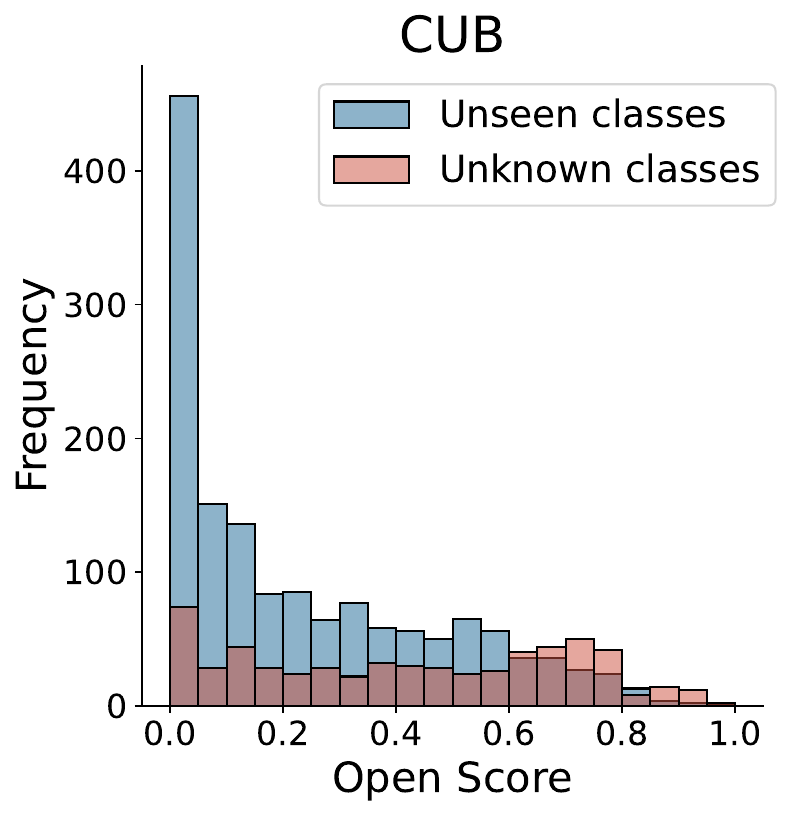}
    \hfill
    \includegraphics[width=0.24\linewidth]{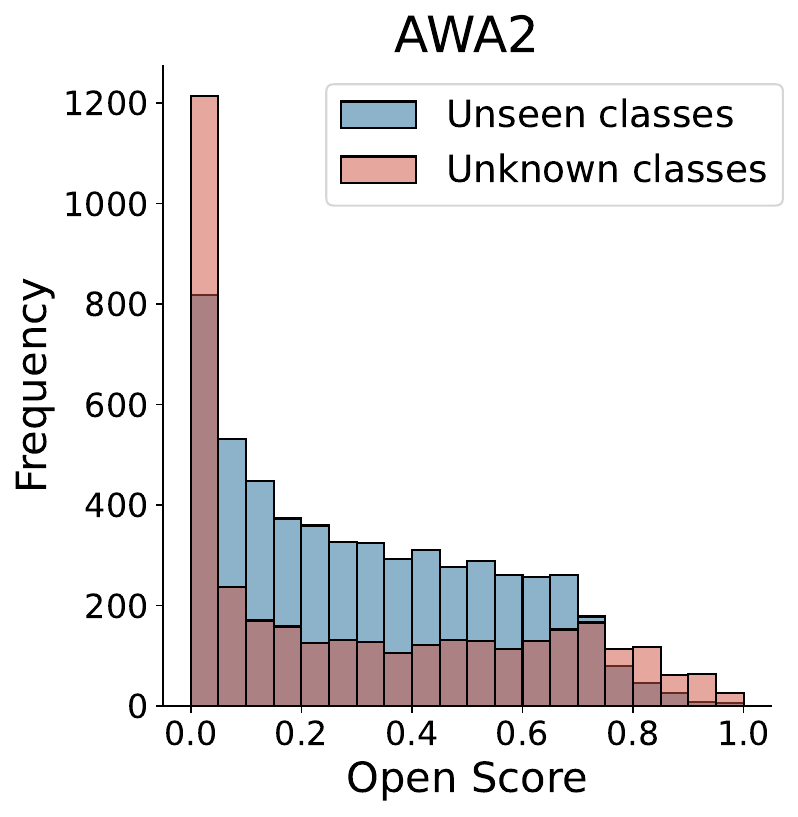}
    \hfill
    \includegraphics[width=0.24\linewidth]{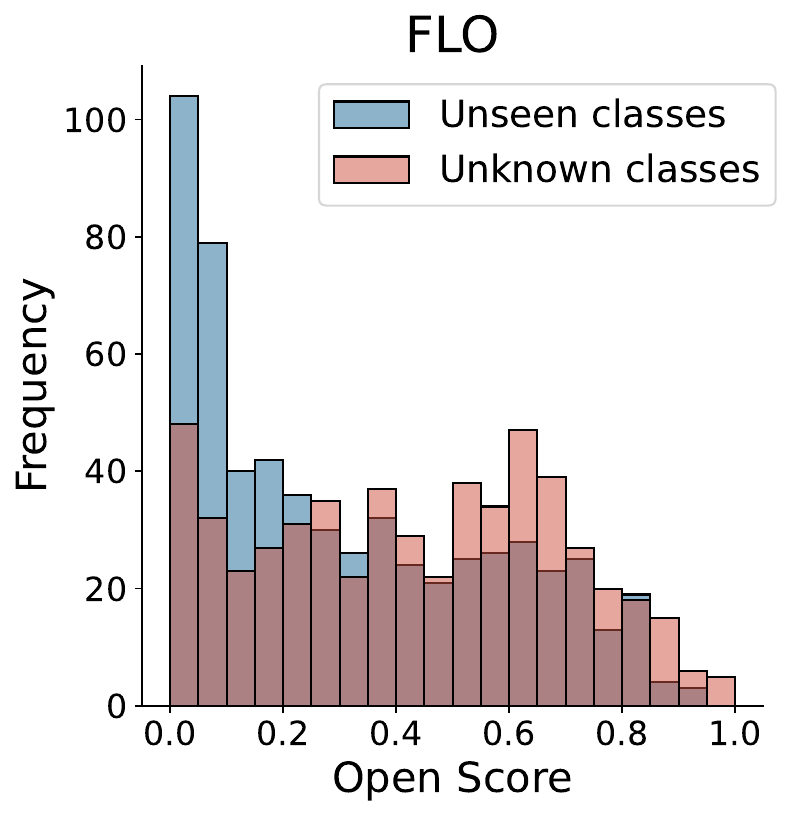}
    \hfill
    \includegraphics[width=0.24\linewidth]{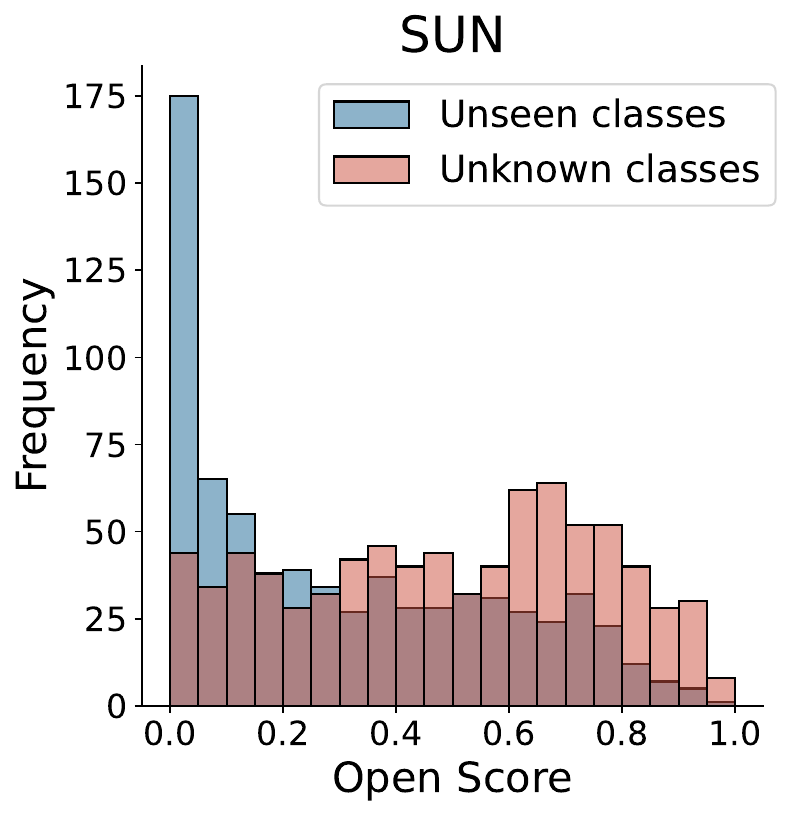}
    \caption{MaxLogit}
  \end{subfigure}
  \begin{subfigure}{1\linewidth}
  \centering
    \includegraphics[width=0.24\linewidth]{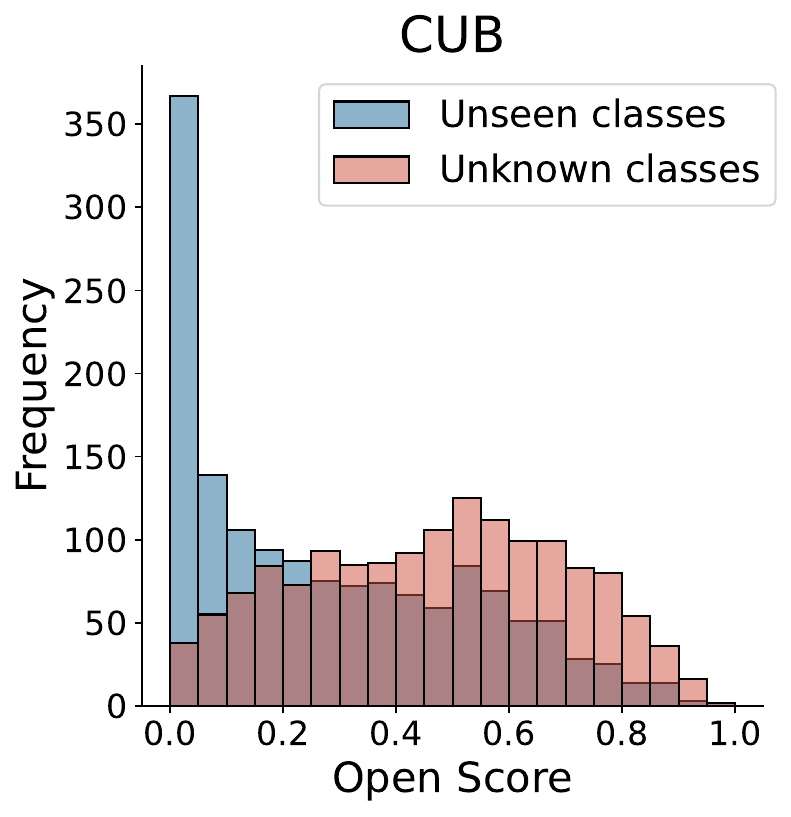}
    \hfill
    \includegraphics[width=0.24\linewidth]{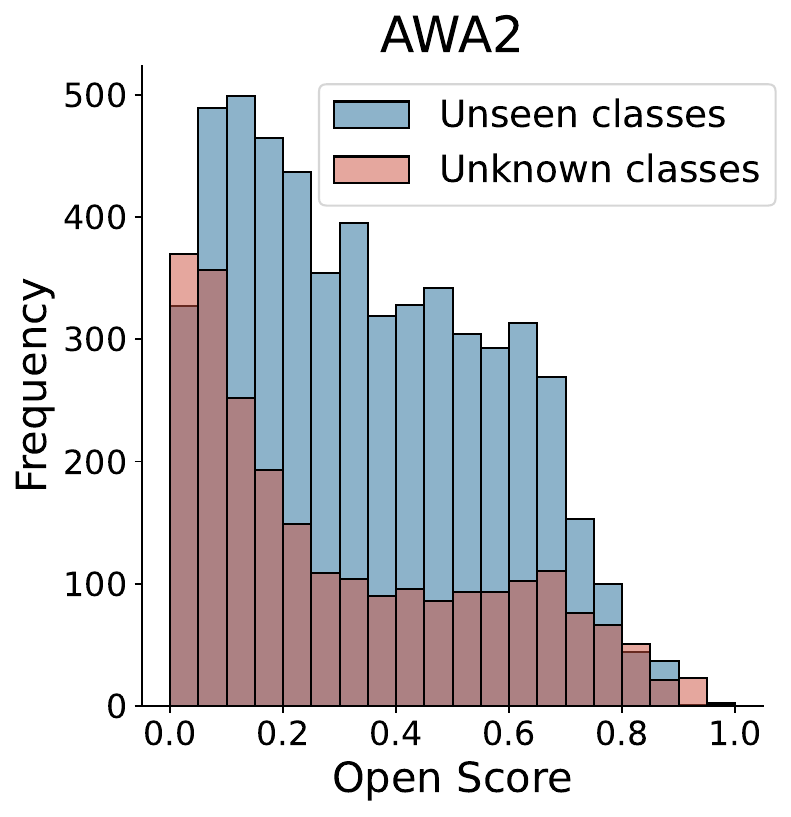}
    \hfill
    \includegraphics[width=0.24\linewidth]{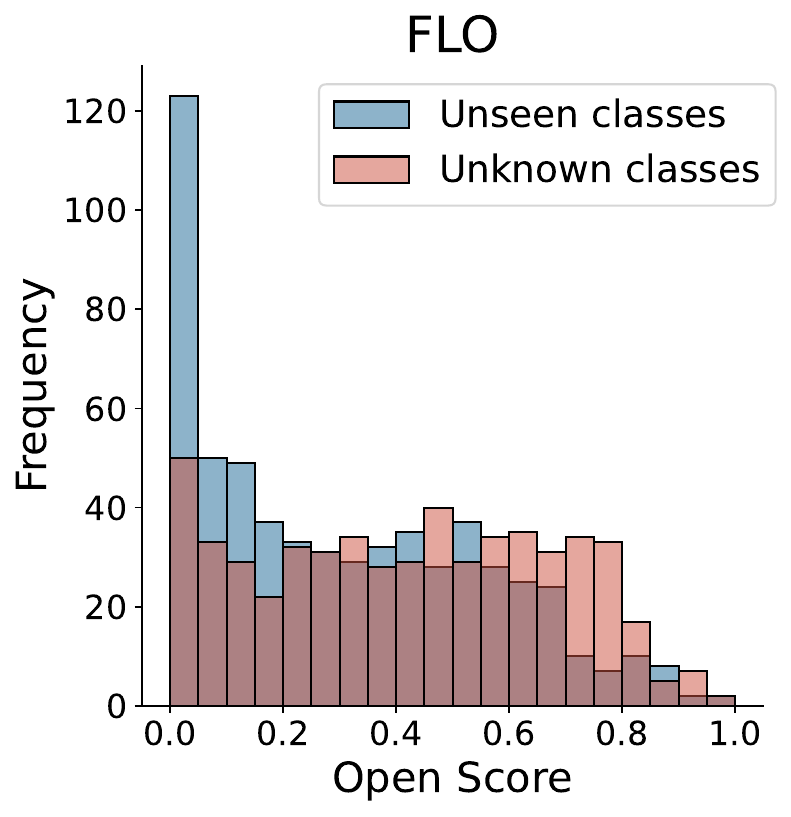}
    \hfill
    \includegraphics[width=0.24\linewidth]{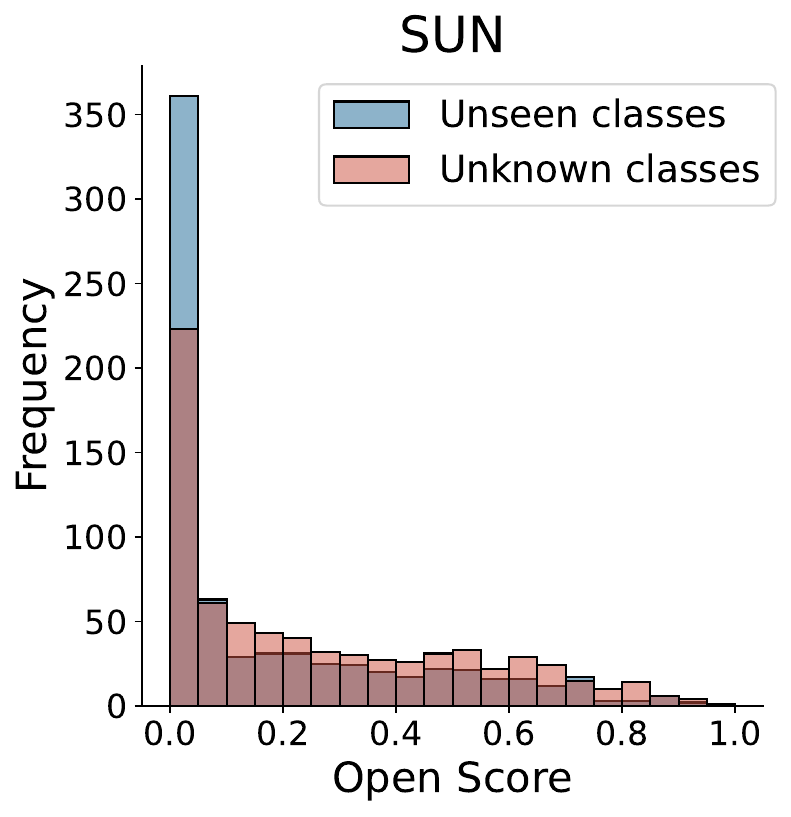}
    \caption{APN}
  \end{subfigure}
  \caption{Distribution of the open scores yielded by combinations of TF-VAEGAN~\cite{narayan2020latent}  and MSP~\cite{hendrycks2017a}, OpenMax~\cite{bendale2016towards}, Placeholder~\cite{DBLP:journals/corr/abs-2103-15086}, Energy~\cite{10.5555/3495724.3497526}, ODIN~\cite{liang2018enhancing}, LogitNorm~\cite{DBLP:conf/icml/WeiXCF0L22}, and
  MaxLogit~\cite{DBLP:conf/icml/HendrycksBMZKMS22}, combination of APN~\cite{NEURIPS2020_fa2431bf} and MSP for the test images of unseen and unknown classes on CUB, AWA2, FLO and SUN datasets.} 
  \label{fig:challenge}
\end{figure*}
Formally, given a training set consisting of $\mathcal{D}^{seen}=\{(\mathbf{x}, y_{\mathbf{x}}, \mathbf{a}_{y}) \mid \mathbf{x} \in \mathcal{X}^{seen}, y_{\mathbf{x}} \in \mathcal{Y}^{seen}, \mathbf{a}_{y} \in \mathcal{A}^{seen}\}$ and $\mathcal{D}^{unseen}=\{(\tilde{y}_{i}, \mathbf{a}_{\tilde{y}_{i}}) \mid \tilde{y}_{i} \in \mathcal{Y}^{unseen}, \mathbf{a}_{\tilde{y}_{i}} \in \mathcal{A}^{unseen}\}$, where $\mathbf{x}$ is an image from the seen class sample set $\mathcal{X}^{seen}$, $y_{\mathbf{x}}$ or $\tilde{y}_{i}$ is a class label, $\mathbf{a}_{y}, \mathbf{a}_{\tilde{y}_{i}} \in \mathbb{R}^M$ are $M$-dimensional class semantic embeddings (\ie, class-level side information), $\mathcal{A}^{seen}$ and $\mathcal{A}^{unseen}$ contain the semantic embeddings of seen and unseen classes respectively. For test images consisting of images of both unseen and unknown classes, \ie, $\mathcal{X}^{unseen} \cup \mathcal{X}^{unknown}$, ZS-OSR aims to learn a model $\phi^{open}$ that maps the images to the class label space $\mathcal{O}=\mathcal{Y}^{unseen} \cup \{'unknown'\}$, where $\mathcal{O}$ includes an '\textit{unknown}' class label in addition to the labels of the unseen classes.

\subsection{The Challenges} \label{sec:challenge}
One of the key challenges in ZS-OSR is the difficulty in distinguishing between the unseen and unknown classes due to the lack of training data for both of them. In typical OSR problems, the most commonly used approach involves conducting OSR first, followed by classification. Nevertheless, in ZS-OSR, none of the currently existing OSR methods have been found to be effective due to the lack of images from the unseen class for training purposes. To address this challenge, a straightforward solution would be to use simple combinations of existing generative ZSL and OSR methods. Specifically, generative ZSL methods can be first applied to generate latent visual features of unseen classes. The ZS-OSR task is then converted to a general OSR task in the visual feature space that contains the features of both seen and unseen classes. OSR methods can then be directly used to learn an open-set classifier using the training set composed of these features to recognize known (unseen) classes while being capable of rejecting unknown classes. Figure \ref{fig:challenge} shows the results of such solutions that uses the widely-used TF-VAEGAN~\cite{narayan2020latent} and MSP~\cite{hendrycks2017a}, OpenMax~\cite{bendale2016towards}, Placeholder~\cite{DBLP:journals/corr/abs-2103-15086}, Energy~\cite{10.5555/3495724.3497526}, ODIN~\cite{liang2018enhancing}, LogitNorm~\cite{DBLP:conf/icml/WeiXCF0L22}, and
  MaxLogit~\cite{DBLP:conf/icml/HendrycksBMZKMS22}, as the generative ZSL method and the OSR methods, respectively, where the open scores are the likelihood scores of being unknown class samples yielded by the open-set classifier, with larger open scores indicating higher likelihood (see Table \ref{tab:main_results} for detailed quantitative results of these methods).

The results show that the method performs poorly in distinguishing between unseen and unknown classes since the open scores of many unseen class samples are large, highly overlapping with that of the unknown class samples. The ineffective performance may be impacted by the fact that the generative ZSL model generates features for the known unseen classes based on the closed-set assumption, without being informed of the possible presence of unknown classes. As a result, these generated unseen features can heavily overlap with that of the unknown classes in the feature space, rendering the subsequent OSR models ineffective. In the next section, we introduce the ASE approach that learns an open-set classifier for the zero-shot setting to address this issue.

\section{The Proposed Approach}\label{sec:method}
\subsection{Overview of Our Approach} \label{ZSL}
\begin{figure}[t]
  \centering
  \setlength{\abovecaptionskip} {0.cm}
  \includegraphics[width=1\linewidth]{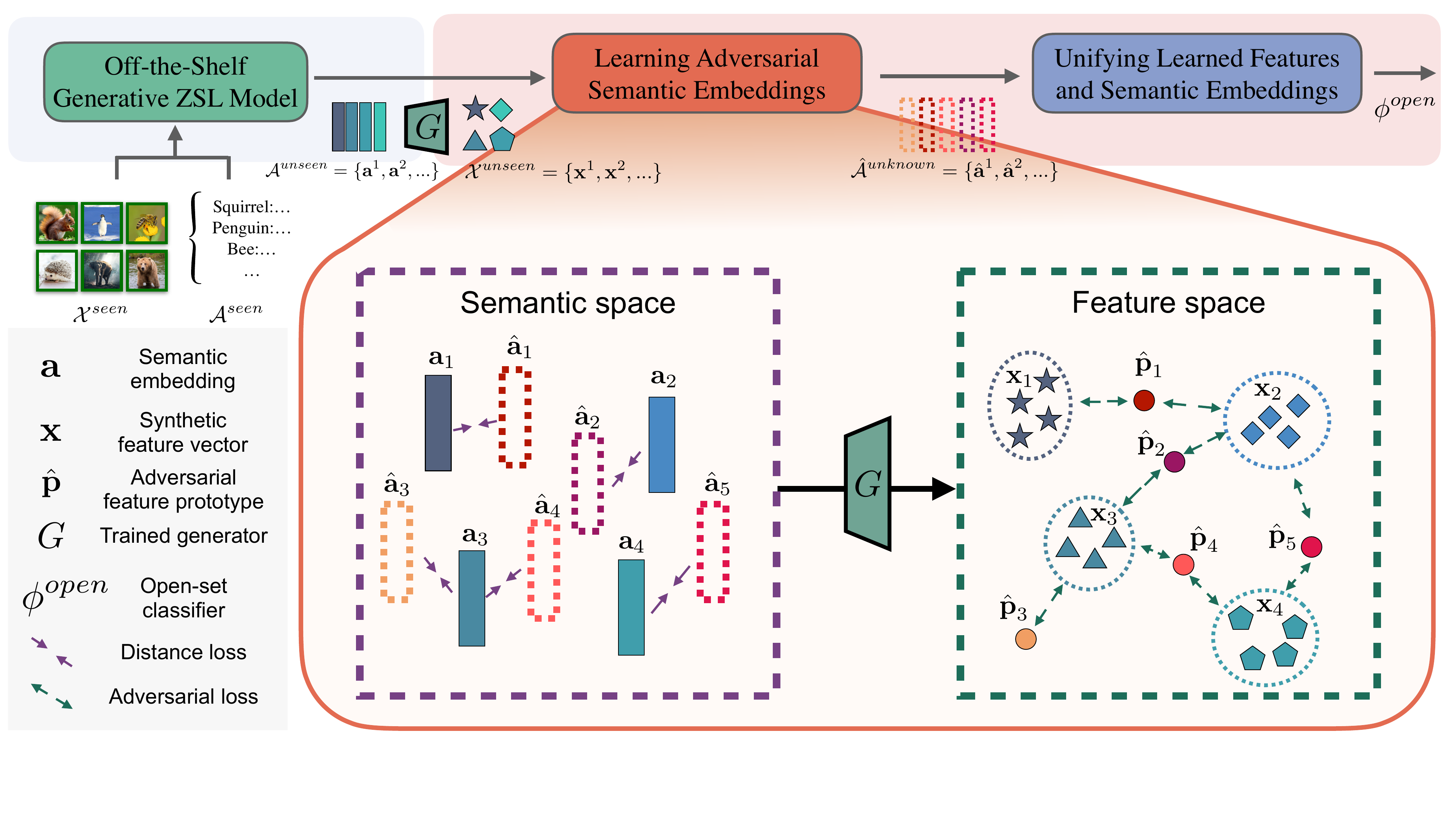}
  \caption{
  The proposed approach ASE. Given a pre-trained generative ZSL model, ASE first uses its generator $G$ to generate the features of unseen classes. It then learns a set of adversarial semantic embeddings of unknown classes so that they are tightly distributed around but separable from the unseen-class embeddings. 
  Lastly ASE uses the unknown-class embeddings to train an unknowns-informed open-set classifier $\phi^{open}$.}
  \label{fig:method_fig}
\end{figure}
Our proposed approach, namely Adversarial Semantic Embeddings (ASE), is a generative framework specifically designed for the ZS-OSR problem. Since ZS-OSR does not provide training samples of unseen and unknown classes, the framework aims to directly generate these samples to train a classifier for this task. Unlike the simple solution in Sec. \ref{sec:challenge} that directly generates the features of unseen classes for the subsequent OSR task, ASE takes a step back and focuses on learning faithful semantic embeddings of unknown classes via an adversarial learning approach before training an unknowns-informed OSR model. 

An overview of ASE is provided in Figure \ref{fig:method_fig}, which consists of three successive components, including \textit{Using off-the-shelf generative ZSL models to generate unseen classes}, \textit{learning adversarial semantic embeddings of unknown classes}, and \textit{unifying these learned features and semantic embeddings to train a $K+1$ open-set classifier} where $K=|\mathcal{Y}^{unseen}|$. The first component is to directly take an existing generative ZSL model to generate the visual features of the unseen classes based on their semantic embeddings. 
The second component is the key novelty of ASE, which is designed to take the trained generator and the closed-set classifier in the ZSL model as input to learn a set of adversarial semantic embeddings of unknown classes, such that the learned semantic embeddings are distributed around the boundary between the known and unknown classes in the semantic space. 
In the third component, these semantic embeddings are subsequently used to generate a set of adversarial feature vectors to represent the samples of the unknown classes, along with the previously generated unseen class samples, to train the $K+1$ classifier for OSR. Below we introduce each component in detail.

\subsection{Using Off-the-Shelf Generative ZSL Models to Generate Unseen-Class Features}\label{sec:unseen}
Training a zero-shot open-set classifier in the visual feature space requires the features of unseen classes. 
Generative ZSL models have demonstrated superior performance in utilizing the relationship between semantic embeddings and image features of the seen classes to generate the unseen-class features. Therefore, existing off-the-shelf generative ZSL models are directly taken to generate these features. Briefly, generative ZSL methods learn a generator network $G$ to generate sample $\tilde{\mathbf{x}}=G(\mathbf{a}, \epsilon)$ conditioned on a Gaussian noise $\epsilon \sim \mathcal{N}(\mathbf{0}, \mathbf{I})$ and a class semantic embedding $\mathbf{a}$. Meanwhile, a discriminator network $D(\mathbf{x}, \mathbf{a})$ is learned by taking an input feature $\mathbf{x}$ and outputs a real value representing the probability that $\mathbf{x}$ is from the real data rather than from the generator network. $G$ and $D$ can be learned by optimizing the following adversarial objective: 
\begin{equation}
  \mathcal{L}=\mathbb{E}[D(\mathbf{x}, \mathbf{a})]-\mathbb{E}[D(\tilde{\mathbf{x}}, \mathbf{a})],
  \label{eq:GAN_loss}
\end{equation}
where $\mathbf{x} \in \mathcal{X}^{unseen}$  is an image from the seen classes and $\mathbf{a} \in \mathcal{A}^{\text{seen}}$ is its class semantic embedding. 
The trained generator generates synthetic features $\tilde{\mathcal{X}}^{unseen} = \{\tilde{\mathbf{x}}_l^{\tilde{y}_i} | \tilde{\mathbf{x}}_l^{\tilde{y}_i}=G(\mathbf{a}_{\tilde{y}_i}, \epsilon_l)\}$ for semantic embedding $\mathbf{a}_{\tilde{y}_i}$ of each class, where $ \mathbf{a}_{\tilde{y}_i} \in \mathcal{A}^{unseen}$. Then we obtain $\tilde{\mathcal{D}}^{unseen}=\{(\tilde{\mathbf{x}}, \tilde{y}_{\tilde{\mathbf{x}}}) \mid \tilde{\mathbf{x}} \in \tilde{\mathcal{X}}^{unseen}, \tilde{y}_{\tilde{\mathbf{x}}} \in \mathcal{Y}^{unseen}\}$, and train a general ZSL (closed-set) classifier $\phi^{closed}$ to classify the test image samples from unseen classes.

Existing state-of-the-art (SOTA) generative ZSL models~\cite{li2019leveraging,xian2019f,narayan2020latent} can be directly used to implement this component. The generator $G$ and the $\phi^{closed}$ classifier (or the generated unseen-class features) are taken as input to adversarially learn the semantic embeddings of unknown classes.

\subsection{Learning Adversarial Semantic Embeddings of Unknown Classes} \label{sec:ase}
ASE is dedicated to learning an unknowns-informed open-set classifier. However, we are given neither semantic embeddings nor image samples of the unknown classes. ASE aims to learn adversarial representations of the unknown classes to train such a classifier. Given the generated features and the training data, the representations of the unknown classes can be adversarially learned in either the semantic embedding space or the visual feature space. 
However, as discussed in Sec. \ref{sec:challenge}, the generated unseen-class features are not faithful enough as they are generated under the closed-set setting. Consequently, directly using these synthetic unseen-class features to generate the unknown-class features may accumulate and/or amplify the closed-set biases in the generator, leading to non-discriminative features for the unknown classes. Such unseen-class and unknown-class features are ineffective in training the open-set classifier (see Table \ref{tab:Ablation}).

Thus, ASE instead learns the unknown-class representations in the semantic space, while enforcing separable unseen-and-unknown representations in the visual feature space. This approach is more plausible since 1) it directly learns the unknown-class semantic embeddings based on the pre-defined embeddings of both seen and unseen classes, while the other approach above is an indirect way that heavily relies on the unstable quality of the generated unseen-class visual features; and 2) it seamlessly leverages the learned relation between the semantic space and the visual feature space in the ZSL models to learn the unknown-class semantic embeddings.

To this end, ASE introduces a class-wise adversarial semantic embedding learning approach to generate a set of semantic embeddings of the unknowns, $\mathcal{A}^{unknown}$. As highlighted in Figure \ref{fig:method_fig}, for each unseen class, ASE generates multiple adversarial semantic embeddings for the unknown classes that are tightly distributed around but separable from the unseen-class embeddings. To achieve this goal, it jointly minimizes a distance loss $\mathcal{L}_{\text{dis}}$ in the embedding space that brings the unknown-class embeddings closer to the unseen-class embeddings, and an adversarial loss $\mathcal{L}_{\text{adv}}$ in the feature space that pulls the corresponding prototypical unknown-class features away from the generated unseen-class features: 
\vspace{-0.1cm}
\begin{equation}
    \mathcal{L}_{\text{ase}}=\mathcal{L}_{\text {adv}}+\beta \mathcal{L}_{\text{dis}},
    \label{eq:total loss}
\end{equation}
\vspace{-0.1cm}
where $\beta$ is a hyper-parameter,
$\mathcal{L}_{\text{dis}}$ is defined as the Euclidean distance between the generated unknown-class embedding $\hat{\mathbf{a}}\in \mathcal{A}^{unknown}$ and the given unseen-class embedding $\tilde{\mathbf{a}}\in \mathcal{A}^{unseen}$:
\vspace{-0.1cm}
\begin{equation}
    \mathcal{L}_{\text{dis}}= \left\|\hat{\mathbf{a}}-\tilde{\mathbf{a}}\right\|_2,
    \label{eq:distance loss}
\end{equation}
\vspace{-0.1cm}
 and $\mathcal{L}_{\text {adv}}$ is defined as a \textit{Helmholtz free energy}-based loss: 
 \vspace{-0.1cm}
\begin{equation}
    \mathcal{L}_{\text{adv}}=T \cdot \log \sum_i^K e^{\phi^{closed}_i(\hat{\mathbf{p}}) / T},
    \label{eq:adversarial loss}
\end{equation}
\vspace{-0.1cm}
where $\phi^{closed}$ is the ZSL (closed-set) classifier obtained from the off-the-shelf ZSL model, and $\hat{\mathbf{p}}=G(\hat{\mathbf{a}}, \epsilon)$ is a generated prototypical unknown-class feature vector corresponding to the adversarial semantic embeddings $\hat{\mathbf{a}}$.
Note that since $\phi^{closed}$ was trained using the unseen-class features and its weight parameters are fixed, the energy score that the unseen-class features receive are consistently low. Thus, Eq. (\ref{eq:adversarial loss}) is designed to encourage high energy scores for unknown-class feature prototypes only.
By minimizing $\mathcal{L}_{\text{ase}}$, the unknown-class and unseen-class class embeddings are close to each other, but they are discriminative from each other; this adversarial relation also applies to the corresponding unknown-class feature prototypes w.r.t. the unseen-class features.
\subsection{Unknowns-Informed ZS-OSR}\label{sec:unifying}
We then train an unknowns-informed ZS-OSR classifier with $K+1$ classes, in which the extra (+1) class is for the `\textit{unknown}' class and it is trained based on the learned unknown-class semantic embeddings.

Specifically, we first utilize the adversarial semantic embeddings $\hat{\mathcal{A}}^{unknown}$ and the trained generator $G$ to obtain a set of unknown-class features $\hat{\mathcal{X}}^{unknown} = \{\hat{\mathbf{x}}_l^{\hat{y}_i}| \hat{\mathbf{x}}_l^{\hat{y}_i}=G(\mathbf{a}_{\hat{y}_i}, \epsilon_l), \mathbf{a}_{\hat{y}_i} \in \hat{\mathcal{A}}^{unknown}, \hat{y}_i \in \mathcal{Y}^{unknown}\}$, where $\mathcal{Y}^{unknown}$ is a set of unknown classes collectively labeled as `\textit{unknown}', resulting in $\tilde{\mathcal{D}}^{unknown} = \{(\hat{\mathbf{x}}, 'unknown') \mid \tilde{\mathbf{x}} \in \hat{\mathcal{X}}^{unknown}\}$. The unknown-class features $\hat{\mathbf{x}}$ are expected to be centered around the unknown-class feature prototypes $\hat{\mathbf{p}}$. The unseen-class and unknown-class features are then combined to form the open-set training data, \ie, $\mathcal{D}^{open}=[\tilde{\mathcal{D}}^{unseen}, \tilde{\mathcal{D}}^{unknown}]$, which is used to train the open-set classifier $\phi^{open}$ by minimizing a standard cross-entropy loss: 
\begin{equation}
    \min _\theta \mathbb{E}_{(\mathbf{x}, y_\mathbf{x}) \sim \mathcal{D}^{open}}\left[-\log \phi^{open}_{y_\mathbf{x}}(\mathbf{x})\right].
    \label{eq:msp}
\end{equation}
During inference, given a test image $\mathbf{x}$,  $\phi^{open}$ yields a softmax score in its $K+1$-th class prediction, which can be directly used as an open score. If the score exceeds a pre-defined threshold, $\mathbf{x}$ is predicted as `\textit{unknown}', and otherwise $\mathbf{x}$ is predicted as the class with the highest logit among the $K$ unseen classes. 
Alternatively, post-hoc OSR methods like MSP~\cite{hendrycks2017a} and ODIN~\cite{liang2018enhancing} can also be applied for obtaining the open score, but the $\phi^{open}$-based open score is generally more effective (shown in Table \ref{tab:Ablation}), and is used by default.

\section{Experiments}

\subsection{Experimental Setup}
\noindent\textbf{Datasets}.
To our knowledge, there are no publicly-available datasets designed for evaluating the performance of ZS-OSR, so we introduce four ZS-OSR datasets adapted from four existing widely-used ZSL datasets: Caltech-UCSD-Birds 200-2011 (CUB)~\cite{WahCUB_200_2011}, Animals with Attributes 2 (AWA2)~\cite{8413121}, FLO~\cite{4756141} and SUN~\cite{6247998}. In particular, we first used the commonly-used seen/unseen class split as in~\cite{8413121} and~\cite{reed2016learning}. However, this data split does not provide the unknown classes. In order to facilitate the ZS-OSR setup on different datasets without loss of generality, for the test data of each dataset, we further randomly take half of the unseen classes as the unknown classes.
In other words, the test data is the same as ZSL datasets but it now contains both unseen and unknown classes; while part of semantic information in the original ZSL training data is removed to create the unknown classes in the test data.
Detailed information of the datasets is presented in Tables \ref{tab:dataset_info} and \ref{tab:dataset_splits}.
\section{Information of Datasets} \label{sec:info_datasets}
\begin{table}
	\begin{footnotesize}
		\centering
		\begin{tabular}{c  c c c c c}
		    \hline
            \hline
            Dataset  & \#images & Dimension & $\mid \textit{KS}\mid $ & $\mid \textit{KU}\mid $ & $\mid \textit{U}\mid $ \\
            \hline
			CUB~\cite{WahCUB_200_2011}  & 11788 & 312 & 150 & 25 & 25 \\
            AWA2~\cite{8413121} & 37322 & 85 & 40 & 5 & 5 \\
            FLO~\cite{4756141} & 8189 & 1024 & 82 & 10 & 10 \\
            SUN~\cite{6247998} & 14340 & 102 & 645 & 36 & 36 \\
			\hline
            \hline
		\end{tabular}
            \vspace{0.3cm}
		\caption{Information on proposed ZS-OSR splits.}
            \vspace{0.2cm}
		\label{tab:dataset_info}
	\end{footnotesize}
  \vspace{-1em}
\end{table}

\begin{table}
	\begin{footnotesize}
		\centering
		\begin{tabular}{c  p{5cm} p{5cm}}
		    \hline
            \hline
            Dataset  & Unseen & Unknown \\
            \hline
            \specialrule{0em}{1pt}{1pt}
            \specialrule{0em}{1pt}{1pt}
            AWA2 & 7, 23, 24, 31, 47 & 9, 30, 34, 41, 50\\
            \specialrule{0em}{1pt}{1pt}
            \specialrule{0em}{1pt}{1pt}
            FLO & 1, 3, 4, 5, 6, 8, 11, 16, 17, 20  &  2, 7, 9, 10, 12, 13, 14, 15, 18, 19\\
            \specialrule{0em}{1pt}{1pt}
            \specialrule{0em}{1pt}{1pt}
            CUB  & 7, 19, 21, 34, 36, 56, 68, 79, 80, 88, 91, 98, 104, 108, 124, 142, 150, 152, 157, 166, 171, 179, 182, 187, 195 &  29, 50, 62, 69, 72, 87, 95, 100, 116, 120, 122, 125, 129, 139, 141, 159, 160, 167, 174, 176, 185, 189, 191, 192, 193\\
            \specialrule{0em}{1pt}{1pt}
            \specialrule{0em}{1pt}{1pt}
            SUN &11, 25, 33, 39, 54, 73, 75, 76, 100, 146, 185, 217, 222, 238, 255, 263, 287, 316, 329, 337, 343, 359, 449, 483, 494, 510, 559, 561, 623, 632, 646, 651, 657, 659, 675, 712 &  4, 24, 58, 86, 96, 104, 113, 125, 131, 139, 153, 159, 197, 246, 247, 260, 299, 354, 380, 382, 421, 424, 426, 441, 472, 509, 518, 530, 581, 636, 680, 682, 696, 711, 713, 716\\
            \specialrule{0em}{1pt}{1pt}
            \specialrule{0em}{1pt}{1pt}
			\hline
            \hline
		\end{tabular}
            \vspace{0.3cm}
		\caption{Proposed splits on ZS-OSR datasets.}
            \vspace{0.2cm}
		\label{tab:dataset_splits}
	\end{footnotesize}
  \vspace{-1em}
\end{table}
\noindent\textbf{Evaluation Metrics}. Following~\cite{bendale2016towards}, we evaluate both closed-set classiﬁcation and open-set detection performance. Classiﬁcation accuracy (Acc) is used to measure the performance of classifying the closed-set samples, \ie, test samples from the unseen classes, while the FPR95 and Area Under ROC Curve (AUROC) are used to measure the performance of detecting open-set samples, \ie, test samples from the unknown classes. All reported Acc, FPR95 and AUROC results are averaged over five independent runs.

\noindent\textbf{Implementation Details of ASE}.
Our ASE approach, outlined in Sec. \ref{sec:method}, relies on an off-the-shelf generative ZSL method, TF-VAEGAN~\cite{narayan2020latent}, to produce feature vectors for unseen classes based on their semantic embeddings. We use ResNet101~\cite{he2016deep} to extract features for $\mathcal{X}$ and conduct a grid search on the validation set~\cite{8413121} to determine the optimal hyperparameter $\beta$ for our model. We generate 50 unknown-class semantic embeddings around each unseen class and produce 1,000 adversarial samples for each unknown-class semantic embedding in the feature space, resulting in $|\tilde{\mathcal{D}}^{unknown}|=1,000 \times |\mathcal{Y}^{unseen}|$ unknown-class samples per dataset. We train the unknowns-informed open-set classifier with a linear classifier featuring one fully connected layer and optimize it with the Adam optimizer using Eq. (\ref{eq:msp}). We maintain the same hyperparameters as in TF-VAEGAN, which we hold fixed throughout the training process of the open-set classifier.

\noindent\textbf{Comparison Baselines}.
Although there are no methods reported to deal with the ZS-OSR problem, SOTA ZSL and OSR methods can be combined to establish some good solutions. Similar to ASE, TF-VAEGAN is used as a SOTA ZSL method here and combined with eight diverse SOTA OSR methods, including \textbf{MSP}~\cite{hendrycks2017a}, \textbf{OpenMax}~\cite{bendale2016towards}, \textbf{ODIN}~\cite{liang2018enhancing}, \textbf{Placeholder}~\cite{DBLP:journals/corr/abs-2103-15086}, \textbf{Energy}~\cite{10.5555/3495724.3497526}, \textbf{LogitNorm}~\cite{DBLP:conf/icml/WeiXCF0L22},
and \textbf{MaxLogit}~\cite{DBLP:conf/icml/HendrycksBMZKMS22},
Since all of these methods are for OSR or OOD detection and do not support ZS-OSR tasks, we adapt them to ZS-OSR using the following method:
1) We first generate the features of the unseen classes using the ZSL generative method TF-VAEGAN, and then 2) we treat the unseen classes as closed-set classes and apply one of these eight OSR methods to recognize the unseen classes while rejecting unknown-class samples. Additionally, we evaluate the performance of a popular non-generative ZSL method \textbf{APN}~\cite{NEURIPS2020_fa2431bf} in conjunction with \textbf{MSP}. 
All the hyperparameters of the baselines are tuned in the same way as ASE on each dataset to have a fair empirical comparison.

\subsection{ZS-OSR Performance}
The ZS-OSR results of ASE and the combined baselines on the four proposed datasets are shown in Table \ref{tab:main_results}. ASE outperforms all eight baseline methods by a significant margin in detecting unknown-class samples and performs comparably well in terms of classification on the unseen-class samples on all four datasets. Details are discussed below.

\noindent\textbf{Superior unknown-class detection}. 
The baseline methods are inconsistent in detecting unknown-class samples on the four datasets, whereas ASE performs consistently well on all of them. ASE significantly improves the AUROC score compared to the best competing baseline on CUB, AWA2, and FLO by margins of 5.22\%, 14.12\%, and 4.64\%, respectively. ASE is also the best performer on SUN, with a relatively marginal improvement. In terms of the FPR95 metric, ASE also demonstrates the best performance across all datasets. Notably, it reduces the FPR95 by 12.42\% compared to the best baseline on AWA2.

\begin{table}[t]
    \centering
    \scalebox{0.7}{
    \begin{tabular}{c| ccc | ccc |  ccc | ccc}
        \hline
        \hline
        \specialrule{0em}{1pt}{1pt}
        \multirow{2}*{\textbf{Method}} & \multicolumn{3}{c|}{\textbf{CUB}} &\multicolumn{3}{c|}{\textbf{AWA2}} &\multicolumn{3}{c|}{\textbf{FLO}} & \multicolumn{3}{c}{\textbf{SUN}}\\
        ~ & Acc $\uparrow$ & FPR95 $\downarrow$ &  AUC $\uparrow$  & Acc $\uparrow$ & FPR95 $\downarrow$ & AUC $\uparrow$ & Acc $\uparrow$ &FPR95 $\downarrow$ &  AUC $\uparrow$& Acc $\uparrow$ & FPR95 $\downarrow$ & AUC $\uparrow$\\
        \specialrule{0em}{1pt}{1pt}
        \hline
        \specialrule{0em}{1pt}{1pt}
        \bf{MSP}~\cite{hendrycks2017a} & 76.33 & 80.96 & 70.08 & 71.56 & 99.61 & 47.51 & 81.41 & 91.33 & 63.74 & \underline{73.33} & 75.00 & \underline{71.63} \\
        \bf{OpenMax}~\cite{bendale2016towards} & 76.41 & 79.54 & \underline{74.98} & 70.33 & 99.43 & 49.86 & 81.34 & 97.16 & 52.80 & 72.94 & 74.03 & 69.75 \\
        \bf{Placeholder}~\cite{DBLP:journals/corr/abs-2103-15086} & 76.11 & 83.25 & 72.43 & 70.63 & 91.43 & 47.25 & {\bf 82.92} & 95.17 & 47.90 & 72.78 & 94.03 & 53.55 \\
        \bf{Energy}~\cite{10.5555/3495724.3497526} & 76.19 & 82.04 & 71.52 & 71.01 & 99.96 & \underline{67.87} & 81.37 & 92.33 & \underline{68.14} & \underline{73.33} & \underline{70.45} & 71.06 \\
        \bf{ODIN}~\cite{liang2018enhancing} & 76.25 & 84.94 & 69.39 & 70.94 & \underline{89.51} & 62.86 & 81.15 & 92.00 & 63.53 & 72.97 & 73.75 & 71.15 \\
            \textbf{LogitNorm}~\cite{DBLP:conf/icml/WeiXCF0L22}, & 75.00 & 72.92 & 73.41 & 69.04 & 99.29 & 45.79 & 78.76 & 92.50 & 67.65 & 66.11 & 76.25 & 65.21 \\
            \textbf{MaxLogit}~\cite{DBLP:conf/icml/HendrycksBMZKMS22} & \underline{76.74} & 82.92 & 71.87 & \underline{71.64} & 99.58 & 46.70 & 80.16 & 91.00 & 62.11 & 72.92 & 74.86 & 69.70 \\
            \textbf{APN}~\cite{NEURIPS2020_fa2431bf} & {\bf 77.41} & \underline{69.82} & 71.82 & 71.41 & 97.66 & 42.71 & 76.53 & \underline{87.62} & 63.31 & 67.50 & 85.00 & 62.58 \\
        \specialrule{0em}{1pt}{1pt}
        \hline
        \specialrule{0em}{1pt}{1pt}
        \bf{ASE} (Ours) & 76.26 &\textbf{68.67} & \textbf{80.20} & \textbf{72.30} &\textbf{77.09} & \textbf {81.99} &\underline{82.44} & \textbf{87.50} & \textbf{72.78} & \textbf{73.61} & \textbf{70.41} & \textbf{72.69} \\
        \specialrule{0em}{1pt}{1pt}
        \hline
        \hline
    \end{tabular}
    }
    \vspace{0.3cm}
    \caption{Main results. Accuracy (\%), FPR95 and AUROC of ASE and baselines on four ZS-OSR datasets.
The best (second-best) results are boldfaced (underlined).}
    \label{tab:main_results}
\end{table}

\noindent\textbf{Maintaining classification accuracy on unseen classes}. 
ASE's superior ability to detect unknown-class samples does not affect its unseen-class classification ability. As seen in Table \ref{tab:main_results}, 
ASE achieves the best overall accuracy performance across the four datasets, as competing methods such as OpenMax and Placeholder have large accuracy drops on certain datasets. 

\begin{figure}
  \centering
  \begin{subfigure}{1\textwidth}
    \centering
    \includegraphics[width=0.24\linewidth]{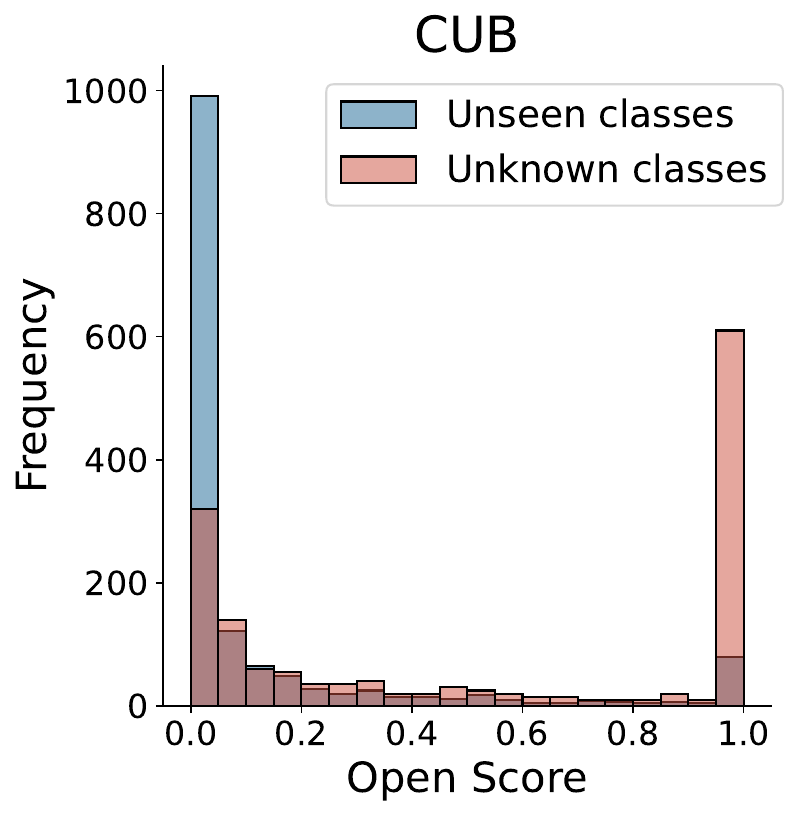}
    \hfill
    \includegraphics[width=0.24\linewidth]{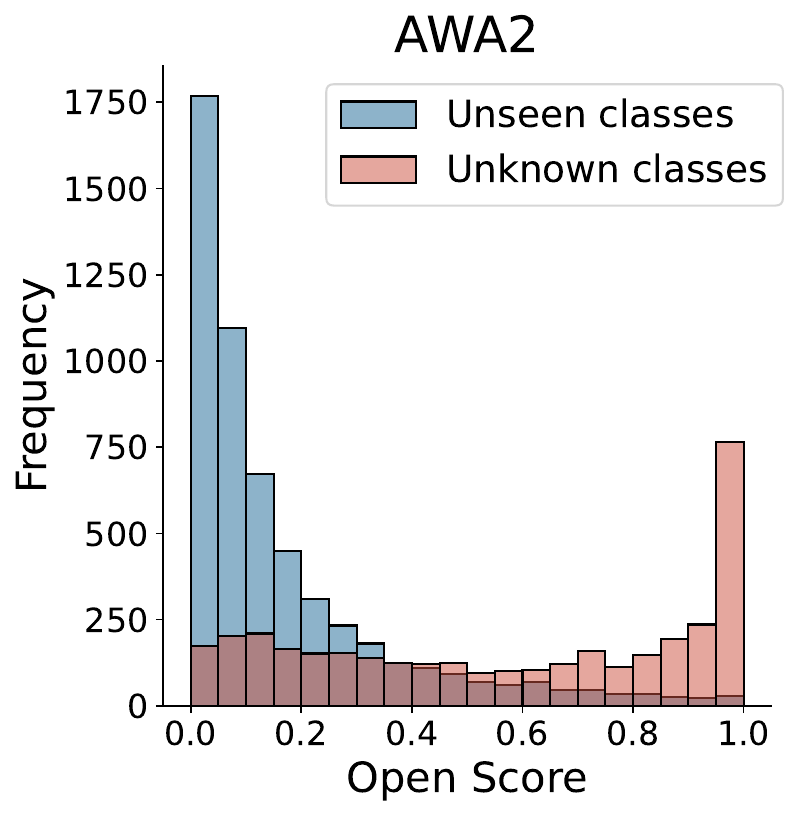}
    \includegraphics[width=0.24\linewidth]{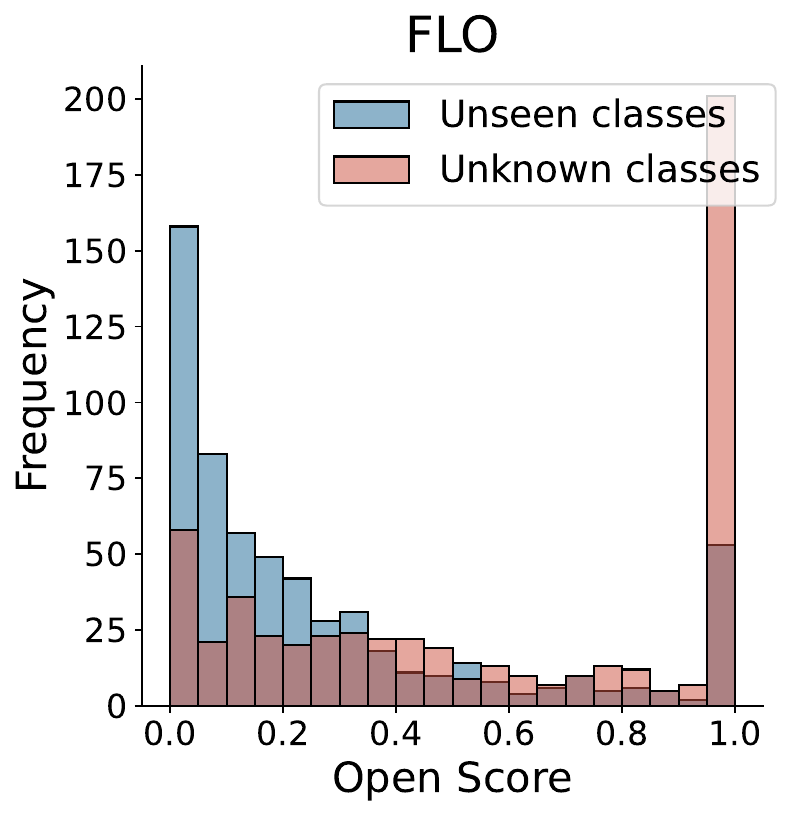}
    \hfill
    \includegraphics[width=0.24\linewidth]{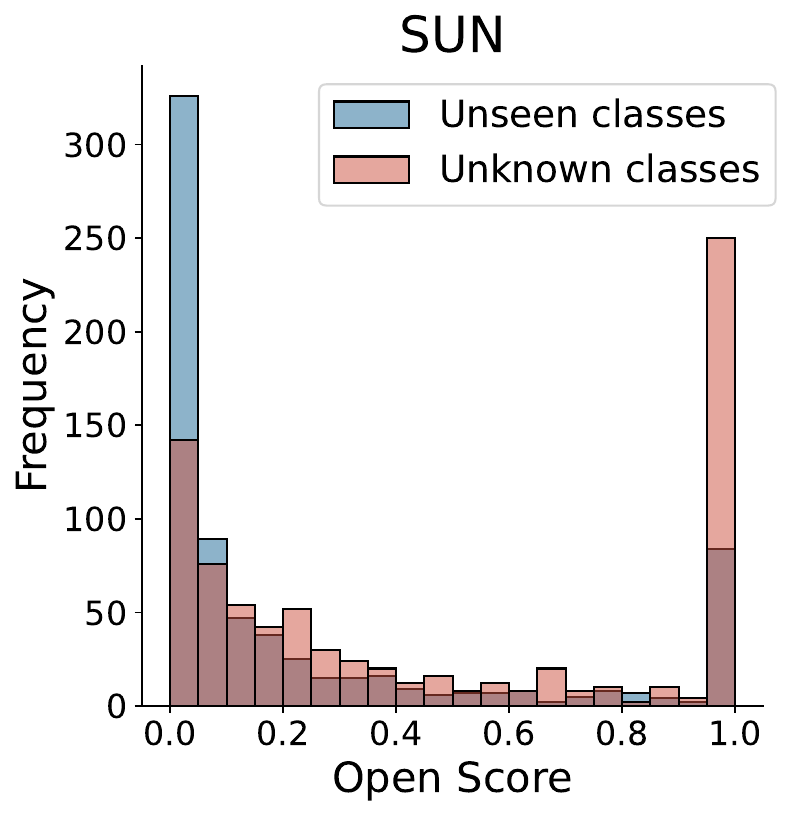}
  \end{subfigure}
  \caption{Distribution of the open scores yielded by ASE.}
  \label{fig:openscore-ase}
\end{figure}

\begin{figure}[t]
  \centering
    \includegraphics[width=0.8\linewidth]{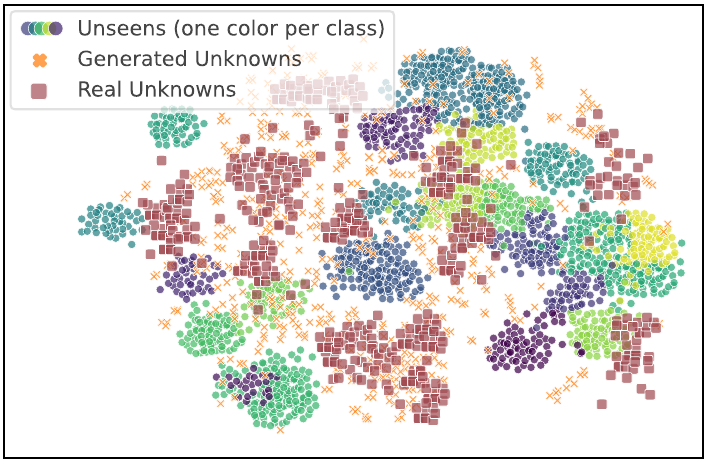}

  \caption{T-SNE visualization of ASE features on CUB.}
  \label{fig:rebuttal_visualization}

\end{figure}

\noindent\textbf{The reasons behind}. As discussed in Sec. \ref{sec:challenge}, the simply combined ZSL-and-OSR solutions fail to produce discriminative open scores for the unseen and unknown class samples, as illustrated in Figure \ref{fig:challenge}. 
In contrast, ASE yields significantly more discriminative open scores for the samples of the unseen and unknown classes, as demonstrated in Figure \ref{fig:openscore-ase}. 
Furthermore, as shown in Figure \ref{fig:rebuttal_visualization}, the unknown samples generated by ASE either lie between unseen and true unknowns or overlap with the real unknown samples, 
suggesting that ASE can effectively leverage the ZSL training data to generate unknown-class representations and distinguish them from the unseen-class samples through the adversarial unknown-class embedding learning in both the semantic and feature spaces.

\subsection{Effectiveness on Data with Varying Openness}
\begin{figure}
  \centering
  \begin{subfigure}{1\linewidth}
    \includegraphics[width=0.39\linewidth]{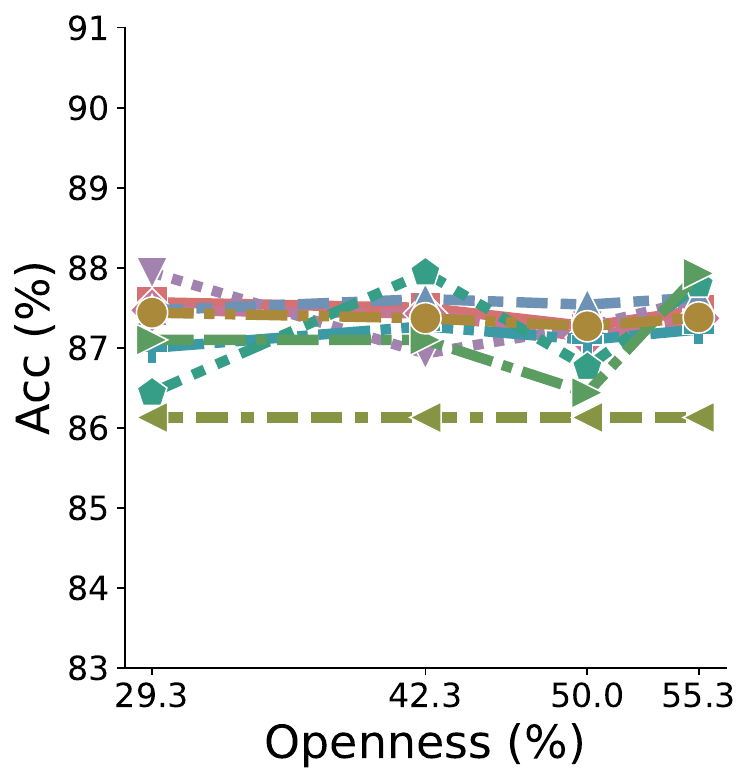}
    \includegraphics[width=0.6\linewidth]{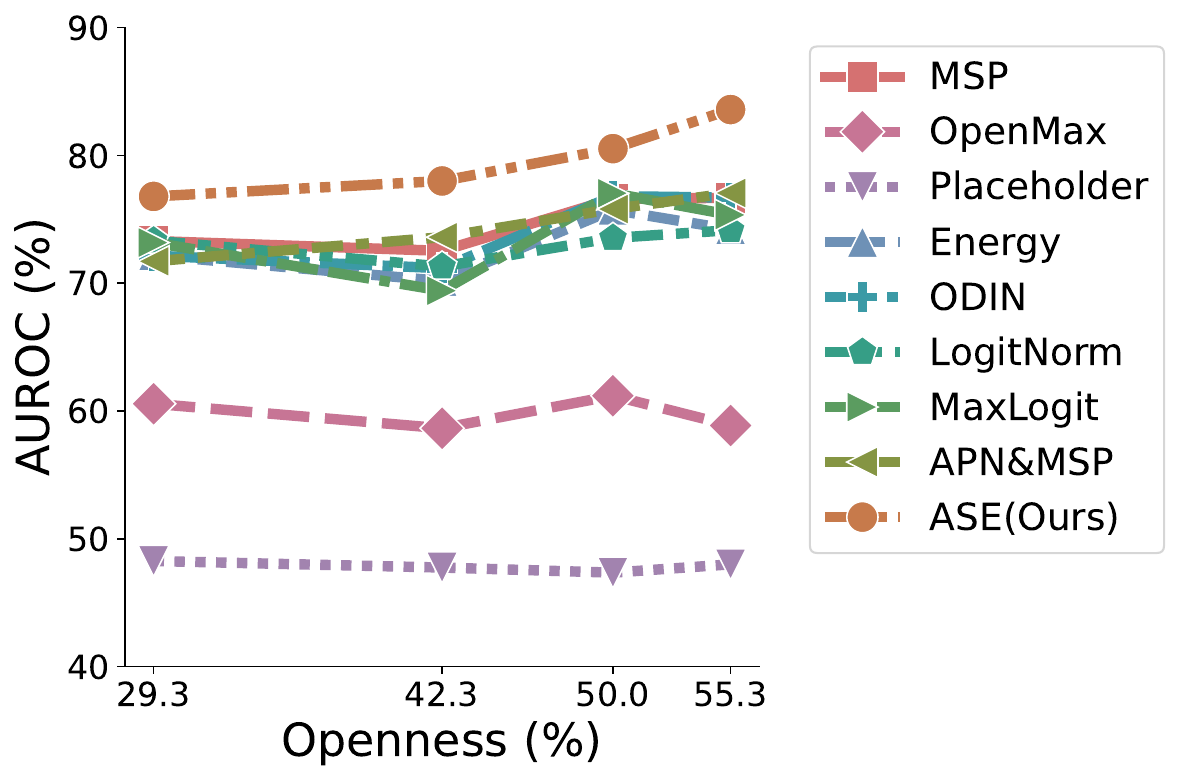}
  \end{subfigure}
  \caption{Performance \wrt openness rates on CUB.}
  \label{fig:openness_fig}
    \vspace{-0.2cm}
\end{figure}

Following the OSR literature~\cite{scheirer2012toward, DBLP:journals/corr/abs-2103-15086, sun2020conditional, perera2020generative}, we conduct experiments on the CUB dataset to examine ZS-OSR performance under varying degrees of openness, defined as $1-\sqrt{\frac{|\mathcal{Y}^{\text {unseen}}|}{|\mathcal{Y}^{\text {unseen}}| + |\mathcal{Y}^{\text {unknown}}|}}$. A larger openness indicates the presence of relatively more unknown classes in the test data. The experiment is focused on the CUB dataset.
In particular, there are 50 unseen classes in CUB under the ZSL setting~\cite{8413121}. We create four ZS-OSR datasets based on CUB by retaining 10 classes as the unseen classes and the respectively remaining 10, 20, 30, and 40 classes as the unknown classes. This results in four datasets with respective openness of 29.3\%, 42.3\%, 50\%, and 55.3\%. The results of ASE and baselines on these four datasets are shown in Figure \ref{fig:openness_fig}. It is clear that
ASE maintains consistently superiority in unknown-class detection and comparably good unseen-class classification across different openness rates, demonstrating strong robustness and stability to the data openness.

\subsection{Ablation Study}\label{subsec:ablation}

Table \ref{tab:Ablation} shows the ablation study results of two main modules in ASE, including $\mathcal{L}_{ase}$ and $\phi^{open}$. We discuss the results in detail below.

\noindent\textbf{Adversarial semantic embedding learning via $\mathcal{L}_{ase}$}. 
We compared the $\mathcal{L}_{ase}$-based adversarial embedding learning module with four alternative methods, including \textit{Mixup}~\cite{zhang2017mixup}, \textit{Uniform Noise}, \textit{Semantic + Noise}, and \textit{Adversarial Features}~\cite{ijcai2017p469,Neal_2018_ECCV}, as described in Section \ref{sec:ase}. Table \ref{tab:Ablation} shows that although some of the simpler methods, such as \textit{Semantic + Noise}, can achieve fairly good performance compared to the baselines in Table \ref{tab:main_results}, our full model ($\mathcal{L}_{ase}$ + $\phi^{open}$) consistently outperforms them in AUROC on all four datasets. While \textit{Adversarial Features} is more effective than the other baselines, it performs rather unstably.
In summary, ASE is consistently the best performer.

\begin{table}[]
	\begin{footnotesize}
		\centering
    \begin{tabular}{c c c c c c}
        \hline
        \hline
        \specialrule{0em}{1pt}{1pt}
        \multirow{2}*{\textbf{Unknowns}} & \multirow{2}*{\textbf{Score}} & \multicolumn{4}{c}{\textbf{AUROC}} \\
        ~ & ~ & CUB  & AWA2 &  FLO & SUN \\
    \hline
    \specialrule{0em}{1pt}{1pt}
    \textit{Mixup}  & $\phi^{open}$ & 60.33 & 25.31 & 46.96 & 57.98\\
    \textit{Uniform Noise}  & $\phi^{open}$ &  63.58 & 46.11 & 50.42 & 57.29\\
    \textit{Semantic + Noise}  & $\phi^{open}$ & 63.74 & 69.02 & 63.22 & 54.77\\
    \specialrule{0em}{1pt}{1pt}
    \specialrule{0em}{1pt}{1pt}
    \textit{Adversarial Features}  & $\phi^{open}$  & 74.92 & 54.70 & \bf{73.19} & 70.75 \\
    \specialrule{0em}{1pt}{1pt}
    \cline{1-6}
    \specialrule{0em}{1pt}{1pt}
    \cellcolor{mygray}\textbf{$\mathcal{L}_{ase}$}  & \cellcolor{mygray}$\phi^{open}$ &  \cellcolor{mygray}{\bf 80.20} & \cellcolor{mygray}{\bf 81.99} & \cellcolor{mygray}{72.78} & \cellcolor{mygray}{72.69}\\
    \specialrule{0em}{1pt}{1pt}
    \cline{1-6}
    \specialrule{0em}{1pt}{1pt}
    \textbf{$\mathcal{L}_{ase}$}  &  MSP  & 71.69 & 75.07 & 69.63 & \bf{76.11}\\
    \textbf{$\mathcal{L}_{ase}$}  &  ODIN  & 72.07 & 74.67 & 66.81 & 71.96\\
    \specialrule{0em}{1pt}{1pt}
    \hline
    \hline
    \end{tabular}
    \vspace{0.3cm}
		\caption{Results of ASE (shaded) and its six variants}
		\label{tab:Ablation}
	\end{footnotesize}
\end{table}

\noindent\textbf{Unknowns-informed open-set classifier via $\phi^{open}$}.
As discussed in Section \ref{sec:unifying}, Post-hoc OSR methods, such as ODIN and MSP, can also be applied to the final classifier trained by ASE. Table \ref{tab:Ablation} shows that ASE-enabled ODIN and MSP methods can largely improve their unknown detection performance compared to the original ODIN and MSP methods (see Table \ref{tab:main_results}), indicating that the final unknowns-informed classifier trained by ASE is more effective in discriminating the unknown samples than the classifier in the original generative ZSL model. However, ASE substantially outperforms both ASE-enabled ODIN and MSP methods in AUROC on CUB, AWA2, and FLO, with a maximal increase of about 9\% on CUB and 7\% on AWA2. Although ASE-enabled MSP obtains the best AUROC on SUN, it performs poorly on other datasets. Overall, since $\phi^{open}$ is end-to-end trained to detect unknown-class samples, it is much more effective than the heuristic methods, ASE-enabled ODIN and MSP.

\subsection{Extending to Generalized ZS-OSR and ZS-OOD Settings}

Our approach focuses on distinguishing unseen and unknown samples in ZS-OSR, but ASE can be extended to function under generalized ZS-OSR settings with a minor modification to Eq. (\ref{eq:distance loss}). ASE generates tightly distributed adversarial semantic embeddings around each of the seen-class and unseen-class semantic embeddings, and can generate unknown-class features and train the $\phi^{open}$ classifier with seen-class, unseen-class, and unknown-class features. 
SOTA generalized ZS model \textbf{GCM-CF}~\cite{yue2021counterfactual} with MSP is included to extend our baselines under this setting. The results are presented in Table \ref{tab:gzsl_results}, where ASE achieves the most effective detection of unknown samples across all four datasets, outperforming the best baseline per dataset by 0.1-4.9\% in AUROC. In terms of FPR95, ASE also maintains a similarly leading position. Although APN surpasses ASE on two datasets in FPR95, ASE substantially outperforms it in both AUROC and FPR on the other datasets.

\begin{table*}[ht]
	\begin{footnotesize}
		\centering
        \scalebox{0.9}{
		\begin{tabular}{c|cc|cc|cc|cc}
		\hline
            \hline
            \specialrule{0em}{1pt}{1pt}
            \multirow{2}*{\textbf{Method}} & \multicolumn{2}{c|}{\textbf{CUB}} &\multicolumn{2}{c|}{\textbf{AWA2}} &\multicolumn{2}{c|}{\textbf{FLO}} & \multicolumn{2}{c}{\textbf{SUN}}\\
			~ & FPR95 $\downarrow$ & AUC $\uparrow$& FPR95 $\downarrow$ & AUC $\uparrow$ & FPR95 $\downarrow$ & AUC $\uparrow$ & FPR95 $\downarrow$ & AUC $\uparrow$\\
            \specialrule{0em}{1pt}{1pt}
            \hline
            \specialrule{0em}{1pt}{1pt}
			 \bf{MSP} & 81.1 & 66.2 & 84.7 & 63.3 & 70.6 & 72.5 & 88.6 & 60.5 \\ 
              \bf{OpenMax} & 92.4 & 55.1 & 95.5 & 59.0 & 81.8 & 66.6 & 94.3 & 48.7 \\ 
              \bf{Placeholder} & 79.3 & \underline{71.0} & 91.9 & 45.9 & 94.8 & 50.4 & 88.0 & 57.4 \\ 
              \bf{Energy} & 89.3 & 64.1 & 89.3 & 50.8 & 81.2 & 52.7 & 91.8 & 59.6 \\ 
              \bf{ODIN} & 85.1 & 67.8 & 78.5 & 68.2 & 80.5 & 75.1 & 89.9 & 60.7 \\ 
              \bf{LogitNorm} & \underline{78.6} & 69.6 & 75.9 & 62.1 & 75.3 & 66.5 & \underline{87.6} & \underline{61.7}\\ 
              \bf{MaxLogit} & 84.2 & 64.8 & 83.7 & 62.6 & 70.2 & 65.4 & 89.1 & 59.5 \\ 
              \bf{APN} & 88.3 & 56.7 & \textbf{69.1} & \underline{73.9} & \bf{54.3} & 71.7 & 90.9 & 56.1 \\ 
             \bf{GCM-CF} & 84.3 & 64.5 & 74.0 & 67.0& \underline{67.3} & \underline{75.7} & 90.2 & 60.6\\
            \specialrule{0em}{1pt}{1pt}
			\hline
            \specialrule{0em}{1pt}{1pt}
			\bf{ASE} (Ours)& \textbf{77.8} & \textbf{75.9} & \underline{73.8} & \textbf{78.7} & 69.4 & \textbf{79.1} & \textbf{87.5} & \textbf{61.8} \\
            \specialrule{0em}{1pt}{1pt}
			\hline
            \hline
		\end{tabular}
        }\vspace{0.3cm}
		\caption{Results under generalized ZS-OSR setting.
		}
		\label{tab:gzsl_results}
	\end{footnotesize}

\end{table*}

\begin{table*}[ht]
	\begin{footnotesize}
		\centering
        \scalebox{1}{
		\begin{tabular}{c|cc|cc|cc}
		\hline
            \hline
            \specialrule{0em}{1pt}{1pt}
            \multirow{2}*{\textbf{Method}} & \multicolumn{2}{c|}{\textbf{CUB-AWA2}} &\multicolumn{2}{c|}{\textbf{CUB-FLO}} &\multicolumn{2}{c}{\textbf{CUB-SUN}} \\
			~ & FPR95 $\downarrow$ & AUC $\uparrow$& FPR95 $\downarrow$ & AUC $\uparrow$& FPR95 $\downarrow$ & AUC $\uparrow$\\
            \specialrule{0em}{1pt}{1pt}
            \hline
            \specialrule{0em}{1pt}{1pt}
			 \bf{MSP}  & 83.0 & 72.6 & 84.2 & 74.2 & 86.4 & 75.0 \\ 
              \bf{OpenMax} & 95.1 & 48.2 & 91.3 & 41.6 & 91.6 & 43.3 \\ 
              \bf{Placeholder} & 79.5 & 70.9 & 78.9 & 72.7 & 81.5 & 72.0 \\ 
              \bf{Energy} & 69.4 & \underline{91.8} & 74.9 & \underline{98.6} & 56.4 & \textbf{99.9} \\ 
              \bf{ODIN} & 91.7 & 69.2 & 78.9 & 62.4 & 83.9 & 70.5 \\ 
              \bf{LogitNorm} & 66.8 & 69.2 & 61.4 & 62.4 & 55.4 & 70.5 \\ 
              \bf{MaxLogit} & 86.7 & 79.4 & 83.1 & 83.7 & 81.4 & 89.0 \\ 
              \bf{APN} & \underline{45.8} & 88.4 & \underline{25.7} & 94.0 & \underline{31.3} & 93.1 \\ 
            \specialrule{0em}{1pt}{1pt}
			\hline
            \specialrule{0em}{1pt}{1pt}
			\bf{ASE} (Ours)& \textbf{4.0} & \textbf{97.2} & \textbf{5.3} & \textbf{99.0} & \textbf{0.3} & \textbf{99.9} \\
            \specialrule{0em}{1pt}{1pt}
			\hline
            \hline
		\end{tabular}
        }\vspace{0.3cm}
		\caption{Results under generalized ZS-OOD setting.
		}
		\label{tab:ood_results}
	\end{footnotesize}
\end{table*}       
While ZS-OSR focuses on the scenario where known classes and unknown classes belong to the same distribution, we believe that there exists scenarios of ZS-OOD (Out-of-Distribution), where we only know the semantic information of the known classes but need to detect images of unknown classes from a different distribution (\eg, samples from a largely different dataset).
To evaluate ASE's performance in the ZS-OOD setting, we randomly select 25 classes from AWA2, FLO, and SUN as unknown classes and combine them with 25 unseen classes of CUB to obtain three ZS-OOD test datasets, CUB-AWA2, CUB-FLO, and CUB-SUN. 
Table \ref{tab:ood_results} shows that considering both AUROC and FPR95 metrics, ASE's performance surpasses all baselines by a significant margin, 
demonstrating the empirical effectiveness of ASE under the ZS-OOD setting.
\section{Conclusions}
This work introduces ZS-OSR, a problem setting which extends ZSL to open-set scenarios, and analyzes the challenge of distinguishing samples of unseen and unknown classes. To promote the development and evaluation of ZS-OSR methods, we build eight baselines that combine SOTA ZSL and OSR models, and establish performance benchmarks by applying them to four ZS-OSR datasets adapted from ZSL datasets. We further propose the ASE approach that learns adversarial semantic embeddings to accurately detect the unknown samples while maintaining preferable classification accuracy of the unseen-class samples. 
Empirical results show that ASE 1) outperforms the baselines on the four datasets in AUROC, 2) performs stably on datasets with varying openness, and 3) can be easily extended to detect the unknown samples under generalized ZS-OSR settings and ZS-OOD settings.

{
\small
\bibliographystyle{elsarticle-num}
\bibliography{ase}
}

\end{document}